\documentclass[10pt]{article}

\usepackage[letterpaper, textwidth=5.5in, textheight=9in,
            top=1in, headheight=12pt, headsep=25pt, footskip=30pt]{geometry}

\usepackage[utf8]{inputenc}
\usepackage[T1]{fontenc}
\usepackage{lmodern}                         
\usepackage{microtype}
\usepackage{graphicx}
\graphicspath{{figures/}}                    
\usepackage{subcaption}
\usepackage{booktabs}
\usepackage{amsmath,amssymb,amsfonts,mathtools,amsthm,bm}
\usepackage{nicefrac}
\usepackage{xcolor}
\usepackage{natbib}                          
\usepackage{hyperref}
\usepackage{url}
\usepackage[capitalize,noabbrev]{cleveref}   
\usepackage{enumitem}
\usepackage{placeins}                        
\usepackage{float}                           
\usepackage{wrapfig}                         
\usepackage{caption}
\theoremstyle{plain}

\theoremstyle{definition}

\theoremstyle{remark}

\title{ProxyKV: Cross-Model Proxy Pruning \\
       for Efficient Long-Context LLM Inference}

\author{%
  Junjie Li \quad Jiong Lou \quad Jie Li \\[6pt]
  Shanghai Jiao Tong University
}

\date{}

\begin{document}
\maketitle

\begin{abstract}
Efficient long-context inference in Large Language Models (LLMs) is severely constrained by the Key-Value (KV) cache memory wall, yet existing pruning methods force a choice between low-latency heuristics that sacrifice precision and high-precision reconstruction methods that incur prohibitive prefilling overhead. To bridge this scoring-cost--accuracy gap, we propose ProxyKV, a cross-model proxy pruning framework that offloads importance scoring to a lightweight intra-family Small-Model Proxy executed asynchronously to the Large-Model Target. To bridge the architectural gap between heterogeneous models, we design the HybridAxialMapper, which disentangles temporal feature extraction from cross-head alignment, together with a Multi-Granularity Hybrid Loss that shifts the learning objective from rigid regression to relative ranking consistency. Across the Llama-3.1, Qwen-2.5, and Qwen-3 families spanning targets from 7B up to 32B parameters on LongBench, SCBench, and RULER, ProxyKV matches KVZip on aggregate (recovering $\sim$$98.7\%$ of its mean accuracy) while delivering up to a $3.21\times$ prefilling speedup on Llama-3.1-8B (dual-GPU; $\sim$$1.5\times$ shared single-GPU) and sustaining the speedup at contexts up to 170k tokens on Qwen-2.5-7B.
\end{abstract}

\section{Introduction}

The capability to process and reason over long-context sequences has become a critical requirement for modern Large Language Models (LLMs), enabling applications ranging from whole-repository code analysis~\citep{jiang2024survey} to complex multi-hop document reasoning~\citep{minaee2024large}. To facilitate efficient autoregressive generation, the Transformer architecture utilizes a Key-Value (KV) cache to store past token activations, effectively eliminating redundant computations~\citep{kwon2023efficient}. However, as the sequence length $N$ increases, this indispensable mechanism introduces a severe memory bottleneck. The linear growth of the KV cache footprint often exceeds the physical memory capacity of individual GPUs, leading to frequent I/O overhead and a prohibitive increase in overall inference latency~\citep{li2024survey}.

KV cache pruning has emerged to mitigate this bottleneck, but existing query-agnostic solutions face a trade-off between scoring efficiency and pruning precision: heuristic methods like H2O~\citep{zhang2023h2o} and SnapKV~\citep{li2024snapkv} are fast but their local observation windows miss global semantic dependencies, while reconstruction-based methods like KVZip~\citep{kim2025kvzip} achieve high precision through global context reconstruction at the cost of a prohibitive secondary prefilling pass on the target model.

\begin{wrapfigure}[15]{r}{0.48\linewidth}
    \vspace{-2.4\baselineskip}
    \centering
    \includegraphics[width=\linewidth]{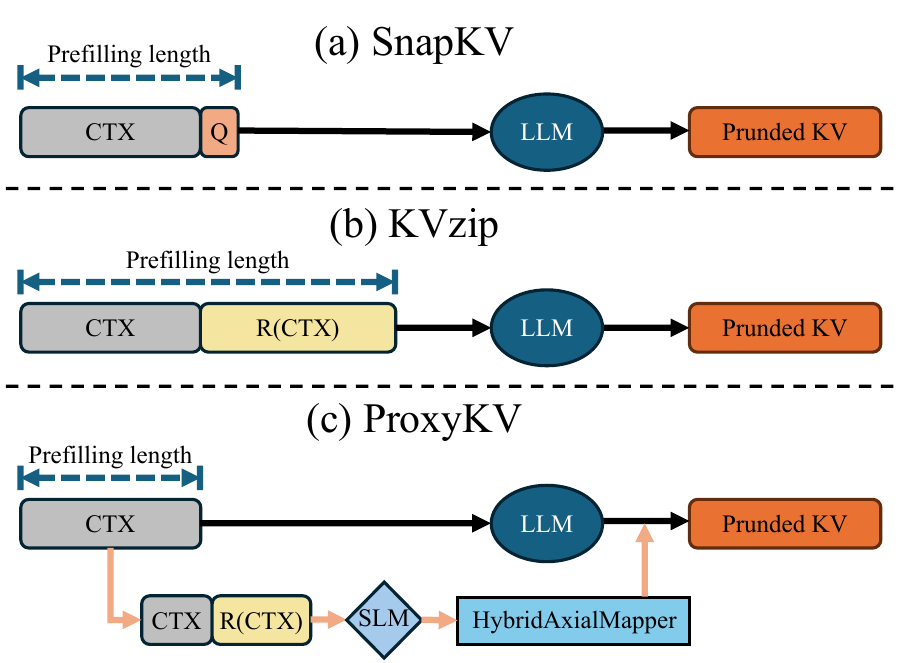}
    \captionsetup{font=normalsize}
    \caption{Three KV-cache pruning paradigms: SnapKV (a) heuristic, KVZip (b) reconstruction, ProxyKV (c) asynchronous proxy.}
    \label{fig:paradigm_compare}
\end{wrapfigure}

As illustrated in \Cref{fig:paradigm_compare}(a, b), the two paradigms anchor opposite ends of the efficiency--precision spectrum. We hypothesize that the high-precision scoring of reconstruction methods can be achieved without the heavy compute on the target, motivated by intra-family attention correlation observed by IAM~\citep{zhao2025iam} and SmallKV~\citep{zhao2025smallkv}; however, existing static head-to-head alignment is too rigid to bridge structural gaps in head count and layer depth.

We propose ProxyKV (\Cref{fig:paradigm_compare}(c)), a cross-model proxy pruning framework that offloads scoring to a lightweight intra-family Small-Model Proxy executed asynchronously to the target's critical path. To handle architectural discrepancies, our HybridAxialMapper disentangles temporal feature extraction from cross-head alignment, and a Multi-Granularity Hybrid Loss shifts the learning objective from rigid value regression to relative ranking consistency---more directly aligned with Top-$K$ pruning quality. Our contributions: (i) an asynchronous proxy-based pruning framework that removes the scoring bottleneck from the target's critical path, delivering up to $3.21\times$ prefilling speedup on Llama-3.1-8B; (ii) the HybridAxialMapper and a five-term Multi-Granularity Hybrid Loss that together bridge structural gaps between heterogeneous models; and (iii) an empirical study across the Llama-3.1, Qwen-2.5, and Qwen-3 families (7B--32B targets) showing ProxyKV recovers $\sim$$98.7\%$ of the KVZip oracle and sustains the speedup at contexts up to 170k tokens.

\section{Related work}

\paragraph{Heuristic and Architectural Pruning.}
Rule-based methods identify non-essential tokens via local patterns: StreamingLLM~\citep{xiao2023efficient} retains attention sinks; H2O~\citep{zhang2023h2o}, Scissorhands~\citep{liu2023scissorhands}, and AhaKV~\citep{gu2025ahakv} use accumulated scores or recent attention patterns. Architectural variants further exploit structure: FastGen~\citep{ge2023model} applies head-wise differentiated policies, PyramidKV~\citep{cai2024pyramidkv} assigns layer-wise budgets, and SnapKV~\citep{li2024snapkv} clusters keys via prompt-end observation windows. ProxyKV instead distills global reconstruction signals, yielding finer-grained pruning that is robust in query-agnostic scenarios.

\paragraph{Learned and Surrogate Pruning.}
Data-driven policies pursue adaptive computation: CoT-Influx~\citep{huang2024fewer} uses RL for coarse-to-fine selection, while GateSkip~\citep{laitenberger2025layers} and LTP~\citep{kim2022learned} introduce differentiable gating. Most recently, KVzap~\citep{jegou2026kvzapfastadaptivefaithful} approximates reconstruction oracles via per-layer surrogates on hidden states; we do not include it as a baseline because (i)~its training cost is substantially higher than ProxyKV's, (ii)~its released implementation adopts a token-selection criterion that differs from the KVzip oracle and is therefore not directly comparable, and (iii)~it does not support evaluation at a fixed retention/compression ratio, which is the standard protocol on LongBench, SCBench, and RULER. Unlike these, which run surrogates within the target's execution, ProxyKV offloads scoring entirely off the critical inference path, with lower training overhead.

\paragraph{Cross-Model KV Cache Alignment.}
Leveraging intra-family correlations, IAM~\citep{zhao2025iam} reuses attention similarity across scales, SmallKV~\citep{zhao2025smallkv} compensates pruning loss with smaller models, and SpeContext~\citep{xu2025specontext} aligns retrieval for speculative prefetch. These rely on rigid head- or layer-wise mappings between fixed pairs; ProxyKV trains a \emph{learnable cross-axial} mapping that jointly models temporal context and head-axis alignment, transferring across architectures with disparate head counts and depths. ProxyKV is also conceptually adjacent to speculative decoding~\citep{kwon2023efficient}, but speculates an \emph{importance-score distribution} for prefill rather than tokens for generation.

\section{System description and problem formulation}
\label{sec:system}

\paragraph{KV cache pruning background.}
Transformer LLM decoding caches the Key/Value tensors of past tokens to avoid redundant compute. As context length $N$ grows the cache becomes a memory wall, motivating pruning that retains only the most informative KV pairs and balances \emph{efficiency} (low scoring latency) against \emph{precision} (long-context reasoning quality).

\paragraph{System overview.}
\label{sec:system_overview}
ProxyKV is a deploy-time pipeline of three components: the Large-Model Target $\mathcal{M}_l$ (owns the prefill critical path and the entire decode phase), an intra-family Small-Model Proxy $\mathcal{M}_s$ that runs asynchronously on a separate execution stream to extract cross-head attention features $\mathbf{X}$, and the HybridAxialMapper $\Phi_\theta$ that maps $\mathbf{X}$ to target-aligned importance scores $\hat{\mathbf{Y}}$. Concretely, $\mathbf{X}$ is the post-softmax attention probability accumulated over the query axis at each proxy layer, i.e., $\mathbf{X}_{b,\ell,h,n} = \sum_{q=1}^{N} \mathrm{Softmax}(\mathbf{Q}\mathbf{K}^{\!\top}\!/\!\sqrt{D})_{b,\ell,h,q,n}$, which compresses a quadratic $N{\times}N$ attention map into a single per-key importance vector of length $N$ per (batch, layer, head). At inference, the input context is dispatched to both models concurrently; $\Phi_\theta$ ingests the proxy features and emits a target-shaped score driving the Top-$K$ pruning mask, after which the target performs decode \emph{without} the proxy on its critical path. The pipeline supports two deployment regimes (\Cref{fig:paradigm_compare}(c)): a dual-GPU regime with parallel target/proxy on separate devices, and a single-GPU regime where they share a device via independent CUDA streams; the proxy KV cache is released at the end of prefill so the proxy-side memory premium is transient (\Cref{fig:memory_timeline}).

\paragraph{Problem formulation.}
\label{sec:problem}
Let $\mathcal{M}_l$ have $L_l$ layers and $H_l$ heads per layer with input length $N$ and per-layer KV tensors $\mathbf{K}, \mathbf{V} \in \mathbb{R}^{B \times H_l \times N \times D}$, and let $\mathcal{M}_s$ have $L_s$ layers and $H_s \ll H_l$ heads per layer. Reconstruction-based oracles (e.g., KVZip) derive the ground-truth attention scores $\mathbf{Y} \in \mathbb{R}^{B \times L_l \times H_l \times N}$ via a secondary teacher-forced prefill on $\mathcal{M}_l$ (\Cref{fig:paradigm_compare}(b)), inflating TTFT. ProxyKV instead approximates $\mathbf{Y}$ via a learnable cross-axial mapping $\hat{\mathbf{Y}} = \Phi_\theta(\mathbf{X}) \approx \mathbf{Y}$, $\mathbf{X} \in \mathbb{R}^{B \times L_s \times H_s \times N}$, and produces a binary mask $\mathbf{M} \in \{0,1\}^{B \times L_l \times H_l \times N}$ by Top-$K$ thresholding $\hat{\mathbf{Y}}$ at retention ratio $\rho \in (0, 1]$. The mapper $\Phi_\theta$ is applied independently to each (proxy-layer, target-layer) pair according to the layer-pairing schedule of \Cref{sec:mapper}, so all per-layer formulations below describe the operation on a single such pair; we drop the explicit $L$ index from $\mathbf{X}, \mathbf{Y}, \mathbf{M}$ in the rest of \Cref{sec:system} and \Cref{sec:loss} to lighten notation. We use \emph{attention scores} for $\mathbf{Y}$ and \emph{importance scores} for $\hat{\mathbf{Y}}$. The training objective is
\begin{equation}
\begin{aligned}
\theta^\star &= \arg\min_{\theta} \;\mathbb{E}_{(\mathbf{X},\mathbf{Y})} \big[ \mathcal{L}\!\left(\Phi_\theta(\mathbf{X}), \mathbf{Y}\right) \big], \\
&\text{s.t.} \;\; t_{\text{score}}(\mathcal{M}_s, \Phi_\theta) \le t_{\text{prefill}}(\mathcal{M}_l), \;\; \mathrm{family}(\mathcal{M}_s) = \mathrm{family}(\mathcal{M}_l),
\end{aligned}
\end{equation}
where $t_{\text{score}}(\mathcal{M}_s, \Phi_\theta) = t_{\text{proxy}}(\mathcal{M}_s) + t_{\Phi_\theta}$ is the total time to run the proxy and the mapper end-to-end, and the async-budget constraint keeps both off the target's critical path (otherwise the speedup vanishes); the intra-family constraint preserves the layer-wise attention correlation that $\Phi_\theta$ exploits. The loss $\mathcal{L}$ is the multi-granularity hybrid loss formalized in \Cref{sec:loss}; the deployment metric is task-accuracy recovery against the KVZip oracle at the same $\rho$.

\section{Methodology}

\begin{figure}[!t]
    \centering
    \includegraphics[width=0.95\linewidth]{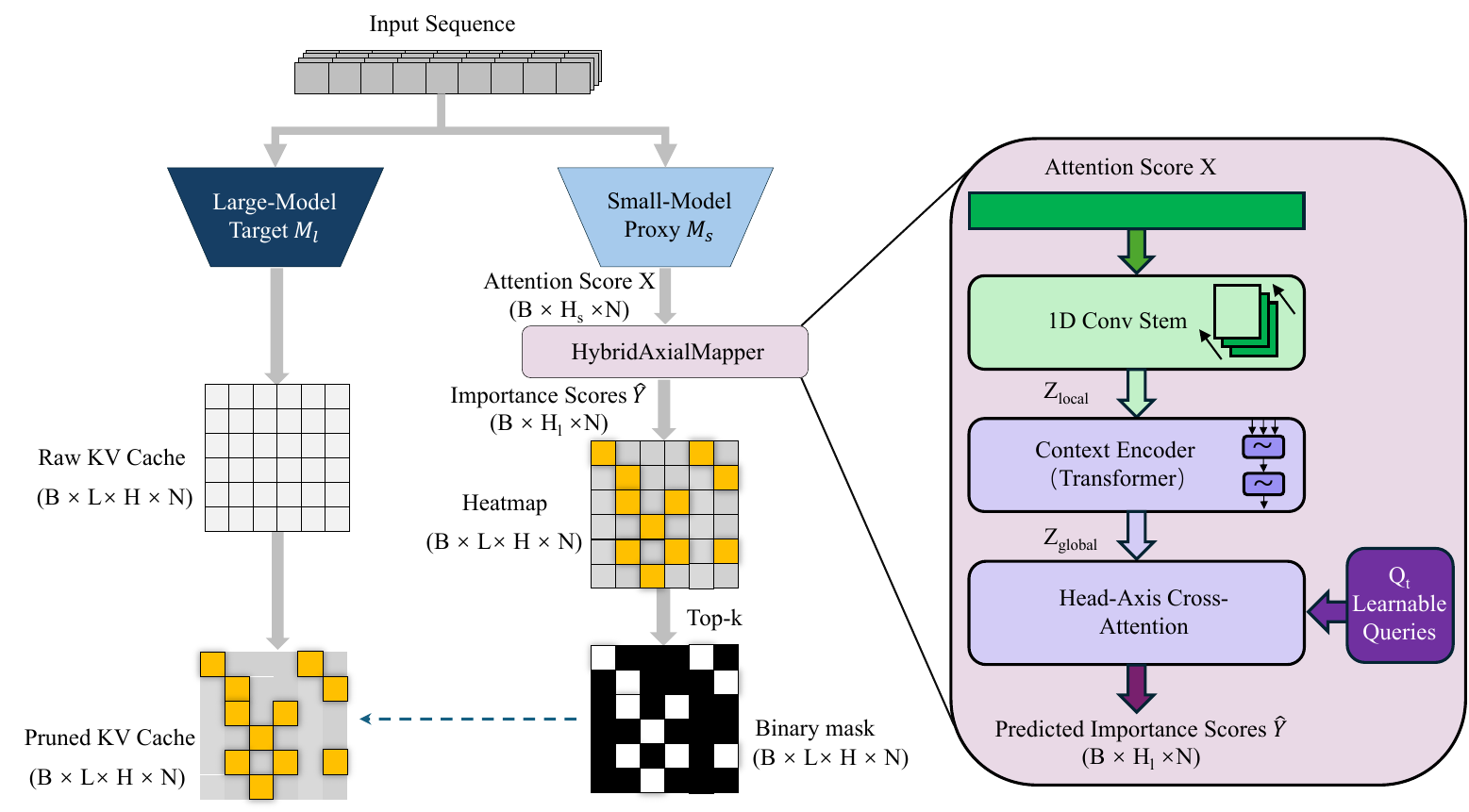}
    \caption{Overview of ProxyKV: an asynchronous Small-Model Proxy $\mathcal{M}_s$ feeds the HybridAxialMapper, which produces target-aligned importance scores $\hat{\mathbf{Y}}$ for the Large-Model Target $\mathcal{M}_l$ without a secondary prefilling pass.}
    \label{fig:methodology}
\end{figure}

Building on the system pipeline of \Cref{sec:system_overview}, the HybridAxialMapper $\Phi_\theta$ transforms the proxy features $\mathbf{X}$ through a three-stage pipeline---(1)~\emph{Temporal Feature Extraction} via a 1D Conv Stem, (2)~\emph{Time-Axis Context Encoder} via a Transformer encoder, (3)~\emph{Head-Axis Cross-Attention} via learnable target queries (\Cref{fig:methodology})---and the predicted scores produce the binary mask $\mathbf{M}$ at the prescribed budget $\mathcal{B}$. The two subsections below detail the mapper architecture and the multi-granularity hybrid loss.

\subsection{HybridAxialMapper architecture}
\label{sec:mapper}

\paragraph{Design rationale.}
Cross-model alignment must reconcile two coupled axes: a \emph{temporal} axis along which token saliency evolves over the sequence, and a \emph{head} axis along which proxy and target attention patterns are partitioned at different granularities ($H_s \ll H_l$). Prior static head-to-head schemes~\citep{zhao2025iam,zhao2025smallkv} entangle these axes by fixing a permutation between proxy and target heads, which prevents transfer to architectures with different head counts or layer depths. Our design instead decomposes the mapping into a temporal stage that produces a head-agnostic latent and a head-axis stage that learns the proxy-to-target aggregation, so the same recipe applies to any intra-family pair. Across the layer axis, we adopt a depth-proportional schedule that pairs target layer $\ell_l \in [1,L_l]$ with proxy layer $\ell_s = \lceil \ell_l \cdot L_s / L_l \rceil$ and runs an independent forward pass of $\Phi_\theta$ for each pair; mapper parameters are shared across all $L_l$ pairs, so $\Phi_\theta$ remains a single network whose forward is invoked $L_l$ times per request and whose output, stacked along the layer axis, recovers the $\mathbf{B \times L_l \times H_l \times N}$ score tensor of \Cref{sec:problem}.

The three-stage pipeline disentangles temporal dependency extraction from head-specific alignment.

\paragraph{Stage 1: temporal feature extraction.}
A 1D Convolutional Stem (two stacked Conv1D layers with $k{=}3$, $p{=}1$, each followed by BatchNorm and GELU) projects the raw proxy features $\mathbf{X}$ into a model-agnostic latent space along the temporal axis:
\begin{equation}
\mathbf{Z}_{local} = \mathrm{GELU}(\mathrm{BN}(\mathrm{Conv1D}(\mathbf{X}))) \in \mathbb{R}^{B \times D_{time} \times N},
\end{equation}
where $D_{time}{=}512$ acts as a bottleneck that normalizes the proxy's head configuration.

\paragraph{Stage 2: time-axis context encoder.}
We transpose $\mathbf{Z}_{local}$ to $\mathbb{R}^{B \times N \times D_{time}}$ (axis swap, not memory reshape), add sinusoidal positional encodings $\mathbf{P}_{pos}$, and process the sequence with a 6-layer Transformer Encoder (8 heads, FFN dim $4 D_{time}$) to produce a globally contextualized $\mathbf{Z}_{global}$. Trained on $2{,}048$-token crops, the encoder is applied at inference via a stride-$1{,}024$ sliding window with overlap averaging, keeping positional indices in-distribution and attention compute at $\mathcal{O}(N)$ even at $170$k tokens.

\paragraph{Stage 3: head-axis cross-attention.}
$\mathbf{Z}_{global}$ is projected and reshaped into latent keys/values $\mathbf{K}, \mathbf{V} \in \mathbb{R}^{(B\cdot N) \times H_s \times D_{head}}$, treating $H_s$ as a synthetic head axis (the Stage~1 $1$D~Conv has already mixed the original proxy heads into the $D_{time}$ bottleneck, so the $H_s$ slots here are learned head-axis tokens rather than the raw proxy heads). A bank of Learnable Target Queries $\mathbf{Q}_{l} \in \mathbb{R}^{H_l \times D_{head}}$ then aggregates these synthetic head tokens into target heads independently per token,
\begin{equation}
\mathbf{O} = \mathrm{Softmax}\!\left(\mathbf{Q}_{l} \mathbf{K}^{\top}/\sqrt{D_{head}}\right) \mathbf{V},
\end{equation}
followed by a learned linear projection $\mathbb{R}^{D_{head}}\!\to\!\mathbb{R}$ to obtain final pruning scores $\hat{\mathbf{Y}} \in \mathbb{R}^{B \times H_l \times N}$. The query bank is the only $H_l$-dependent parameter, so the mapper trivially scales to arbitrary target head counts.

\paragraph{Parameter and FLOP budget.}
The complete mapper has $\sim$$15$M trainable parameters across the three stages---roughly $0.5$M in the convolutional stem, $13$M in the six-layer transformer encoder, and $1.5$M in the head-axis cross-attention block (queries, KV projections, output linear)---two-to-three orders of magnitude smaller than any target we evaluate. Forward FLOPs scale as $\mathcal{O}(N \cdot D_{time}^{2})$ from the encoder and $\mathcal{O}(N \cdot H_l \cdot H_s \cdot D_{head})$ from the cross-attention, so the dominant term is linear in sequence length and independent of the target's hidden width; this is what bounds the mapper's wall-clock share to a single-digit percentage of prefill in \Cref{fig:latency_mapper} and ensures the share stays bounded as targets grow from 7B to 32B.

\subsection{Multi-granularity hybrid loss}
\label{sec:loss}

\paragraph{Why ranking-consistent loss.}
The deployment metric is Top-$K$ retention quality, not pointwise score reconstruction; under sparse, long-tailed attention distributions a small absolute regression error at non-Top-$K$ tokens can flip the binary mask, so MSE alone trains the mapper to imitate magnitudes the target itself never uses. We therefore decompose the objective into one term that supervises the binary decision boundary at multiple retention ratios and four auxiliaries that supervise complementary structural facets of the score distribution.

We minimize a composite loss $\mathcal{L}_{total} = \sum_{k\in\mathcal{K}} \lambda_k \mathcal{L}_k$ with $\mathcal{K}=\{bin, mse, fine, global, cos\}$, centered on a Multi-Ratio Binary term and complemented by auxiliaries covering numerical magnitude, intra-Top-$K$ order, signal--noise boundary, and directional drift.

\paragraph{Multi-ratio binary $\mathcal{L}_{bin}$ and value-weighted MSE $\mathcal{L}_{mse}$.}
The mapper's final linear layer emits a single scalar logit per (head, token); this logit feeds two parallel heads with different output activations: a sigmoid head $\sigma(\hat{\mathbf{Y}})$ that yields a probability for the binary objective $\mathcal{L}_{bin}$, and a magnitude head $\sigma(\hat{\mathbf{Y}})\cdot s_{\max}$ that rescales the same logit to the target's attention-score range for the regression objective $\mathcal{L}_{mse}$ (where $s_{\max} = \max_{ij}\mathbf{Y}_{ij}$ is the per-batch ground-truth maximum). This shared-logit, dual-head design eliminates the domain conflict between BCE (which pushes the logit to $\pm\infty$ for confident decisions) and MSE (which would otherwise pull the same logit to a finite magnitude $\geq 0$). For retention ratios $\mathcal{R}=\{0.05, 0.1, \ldots, 0.5\}$ we materialize binary ground-truth masks $\mathbf{M}_r^\ast$ from the target's Top-$K$ threshold and apply a power-law weighted BCE that emphasizes aggressive compression; $\mathcal{L}_{mse}$ is a value-weighted MSE with weights $w_{ij}=(\mathbf{Y}_{ij}+\epsilon)^{1.5}$ ($\epsilon{=}0.1$) so salient tokens dominate the magnitude error:
\begin{equation}
  \mathcal{L}_{bin} = \!\!\sum_{r \in \mathcal{R}}\! \left(\!\frac{r_{min}}{r}\!\right)^{\!\!\gamma} \!\mathrm{BCE}\!\left(\sigma(\hat{\mathbf{Y}}), \mathbf{M}_r^*\right),
  \quad
  \mathcal{L}_{mse} = \frac{1}{|\Omega|} \!\sum_{(i,j) \in \Omega}\! w_{ij}\, \big( \sigma(\hat{\mathbf{Y}}_{ij})\cdot s_{\max} \!-\! \mathbf{Y}_{ij} \big)^{2},
\end{equation}
with $\gamma{=}1.0$ and $r_{min}{=}0.05$.

\paragraph{Intra-Top-$K$ rank $\mathcal{L}_{fine}$, mass-aware global rank $\mathcal{L}_{global}$, cosine $\mathcal{L}_{cos}$.}
$\mathcal{L}_{fine}$ enforces relative order \emph{within} the target's Top-$K$ set via a softplus pairwise loss weighted by $|\mathbf{Y}_i \!-\! \mathbf{Y}_j|$; $\mathcal{L}_{global}$ is a margin-based pairwise loss over (positive $i\!\in\!\mathrm{TopK}$, negative $j\!\notin\!\mathrm{TopK}$) samples with quality-aware weights $w_{ij}^{global}=\mathrm{clip}(1+|y_{i,\mathrm{norm}} - y_{j,\mathrm{norm}}|,1,5)$ on the min-max-normalised scores $y_{i,\mathrm{norm}}$, upweighting well-separated pairs that are most informative for the signal-to-noise boundary; $\mathcal{L}_{cos}$ aligns the global orientation of the predicted score vector to the target:
\begin{equation}
  \mathcal{L}_{fine} = \mathbb{E}_{i,j\in\mathrm{TopK}}\!\left[\bar{w}_{ij} \ln\!\!\left(1 + e^{-\mathrm{sgn}(\mathbf{Y}_i-\mathbf{Y}_j)\,(\hat{\mathbf{Y}}_i-\hat{\mathbf{Y}}_j)}\right)\right],
  \quad
  \mathcal{L}_{cos} = 1 - \frac{\hat{\mathbf{Y}} \!\cdot\! \mathbf{Y}}{\|\hat{\mathbf{Y}}\|\, \|\mathbf{Y}\|},
\end{equation}
\begin{equation}
  \mathcal{L}_{global} = \mathbb{E}_{i\in\mathrm{TopK},\, j\notin\mathrm{TopK}}\!\left[w_{ij}^{global}\, \big[\,1 - (\hat{\mathbf{Y}}_i - \hat{\mathbf{Y}}_j)\,\big]_{+}\right],
\end{equation}
with $\bar{w}_{ij}=|\mathbf{Y}_i\!-\!\mathbf{Y}_j|/\mathbb{E}[|\mathbf{Y}_i\!-\!\mathbf{Y}_j|]$ and $[\,\cdot\,]_+ = \max(0, \cdot)$. Pairs with score differences below $1\%$ of the max are filtered for numerical stability. We set $\lambda_{mse}{=}20$, $\lambda_{bin}{=}10$, $\lambda_{fine}{=}3$, $\lambda_{global}{=}2$, $\lambda_{cos}{=}0.5$ to balance gradient magnitudes. The five terms supervise complementary failure modes ($\mathcal{L}_{mse}$ magnitude, $\mathcal{L}_{fine}$ near-tied Top-$K$ order at $\rho\!\le\!0.2$, $\mathcal{L}_{global}$ signal-to-noise margin, $\mathcal{L}_{cos}$ anti-drift) and the LOO study in \Cref{sec:ablation} confirms they own nearly disjoint task families at $\rho{=}0.1$, justifying targeted auxiliaries over interchangeable regularizers.

\subsection{Asynchronous scheduling and inference-time pipeline}
\label{sec:async}

\paragraph{Stream-level overlap.}
At inference, $\mathcal{M}_l$ and $\mathcal{M}_s$ are dispatched to two independent CUDA streams (separate GPUs in the dual-GPU regime, distinct streams on the same device in the single-GPU regime); $\Phi_\theta$ is enqueued on the proxy stream as soon as $\mathbf{X}$ is ready, and the resulting $\hat{\mathbf{Y}}$ is consumed by a Top-$K$ kernel on the target stream that materializes $\mathbf{M}$ in place. The async-budget $t_{\text{proxy}} + t_{\Phi_\theta} \le t_{\text{prefill}}(\mathcal{M}_l)$ holds empirically by an order of magnitude on every pair we evaluate; once $\mathbf{M}$ is applied the proxy KV cache is released and decode runs on the target alone (\Cref{fig:memory_timeline}).

\paragraph{Training procedure.}
$\Phi_\theta$ is trained offline against a frozen oracle: we run one teacher-forced KVZip pass per input to cache $(\mathbf{X},\mathbf{Y})$, after which each gradient step costs a single mapper forward/backward. We use AdamW (lr $2{\times}10^{-4}$, weight decay $10^{-4}$) with a $1{,}000$-step linear warmup followed by \texttt{ReduceLROnPlateau} (factor $0.5$, patience $3$), batch size $8$ (effective $12$ across $4$ GPUs for the Qwen-3-32B mapper), sliding-window crops of $2{,}048$ tokens (stride $1{,}024$), and grad-clip $1.0$; loss coefficients are held fixed. Each mapper trains for $30$ epochs without per-pair tuning; per-pair wall-clock and full hyperparameters appear in \Cref{appendix:training_settings}.

\section{Experiments}

We evaluate ProxyKV along three axes: (i) pruning accuracy across diverse benchmarks, (ii) prefilling latency reduction, and (iii) ablations of the mapper and loss design. All experiments use PyTorch on two NVIDIA RTX PRO 6000 GPUs (the 32B target additionally uses a multi-GPU deployment for KVZip).

\subsection{Experiment setup}

\paragraph{Model families.}
We use three intra-family Target--Proxy pairs from 7B to 32B targets, training a dedicated HybridAxialMapper per pair: \textbf{Qwen-2.5}~\citep{yang2025qwen2} (7B / 1.5B), \textbf{Llama-3.x}~\citep{grattafiori2024llama3herdmodels} (Llama-3.1-8B / Llama-3.2-1B; treated as one intra-family pair), and \textbf{Qwen-3}~\citep{yang2025qwen3} (32B / 4B), which stresses ProxyKV at a $\sim$$8\times$ target/proxy size ratio---the largest gap we evaluate.

\paragraph{Baselines.}
We compare against three representative methods. \textbf{SnapKV}~\citep{li2024snapkv} is a training-free heuristic that selects KV pairs from prompt-end observation windows (enhanced variant with global non-uniform budget). \textbf{KVZip+IAM}~\citep{zhao2025iam} adapts IAM's static layer-to-layer alignment to KVZip, isolating our learnable mapper's contribution. \textbf{KVZip}~\citep{kim2025kvzip} is the reconstruction-based oracle, running a full secondary prefilling pass on the target to obtain precise attention scores.

\paragraph{Data and metrics.}
The mapper $\Phi_\theta$ is trained on GSM8K~\citep{cobbe2021training}, SQuAD~\citep{rajpurkar2016squad}, NIAH~\citep{kamradt2023needle}, and the SCBench~\citep{li2024scbench} subsets \textit{QA ENG}, \textit{Summary}, \textit{KV}, \textit{Many Shot}, \textit{Choice Eng}, and \textit{Prefix Suffix}; SCBench.RepoQA is held out for zero-shot evaluation, and we additionally evaluate on the full LongBench~\citep{bai2024longbench} suite. We report Accuracy, Exact Match (EM), and F1 across retention ratios $\rho \in \{0.1, 0.2, \ldots, 0.9\}$, where smaller $\rho$ is more aggressive pruning.

\subsection{Main results}

\raggedbottom
\setlength{\intextsep}{4pt plus 0pt minus 0pt}
\setlength{\textfloatsep}{4pt plus 0pt minus 0pt}
\setlength{\floatsep}{4pt plus 0pt minus 0pt}
\makeatletter
\setlength{\@fpsep}{8pt plus 0fil minus 0fil}      
\setlength{\@fptop}{0pt plus 0fil minus 0fil}      
\setlength{\@fpbot}{0pt plus 1fil minus 0fil}      
\makeatother

\begin{figure}[!ht]
    \centering
    \includegraphics[width=\linewidth]{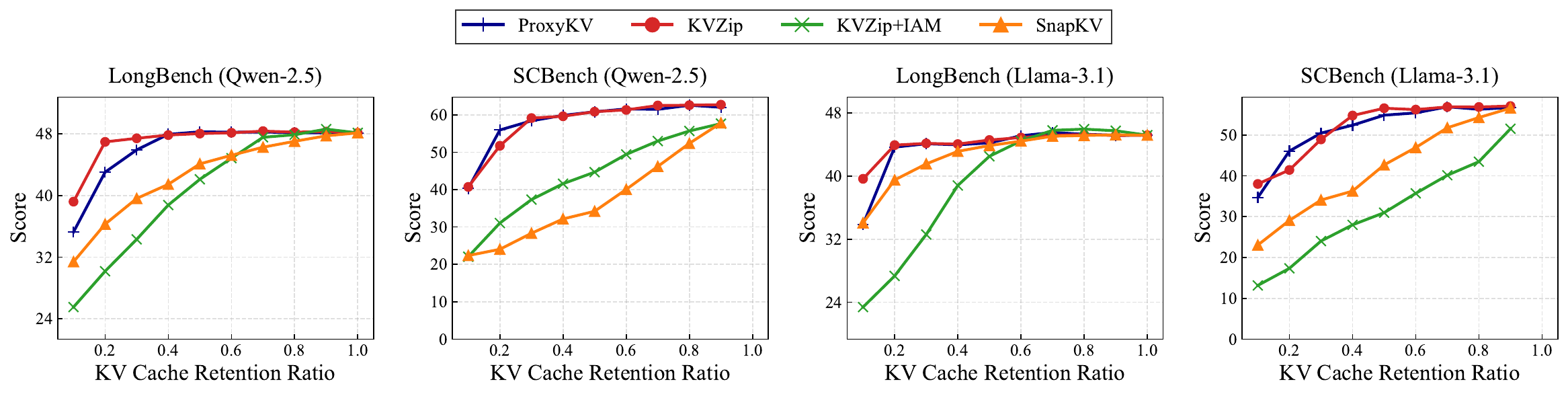}
    \caption{Aggregate accuracy on LongBench and SCBench for the Llama-3.1 and Qwen-2.5 families. ProxyKV tracks the KVZip oracle within $\sim$$1.5$ pp at $\rho \geq 0.5$ (the gap widens to $\sim$$5$ pp at $\rho \leq 0.2$, where pruning bites hardest) and outperforms heuristic SnapKV on SCBench.}
    \label{fig:main_results}
\end{figure}

\textbf{Competitive performance across model families.}
\textit{ProxyKV recovers $\sim$$98.7\%$ of the KVZip oracle across all benchmarks and sparsity levels.} As shown in \Cref{fig:main_results}, ProxyKV (blue) tracks KVZip (red) on every $\rho$, and its margin over KVZip+IAM (green) confirms that the learnable HybridAxialMapper is essential for bridging the head-count / layer-depth mismatches that static alignment cannot resolve. SnapKV (orange) declines on SCBench because its local observation windows miss long-range dependencies; it remains competitive only on tasks with strong syntactic or repetitive patterns (\textit{RepoBench-P}, \textit{TREC}).

\begin{figure}[!ht]
    \centering
    \begin{minipage}[c]{0.5\linewidth}
        \includegraphics[width=\linewidth]{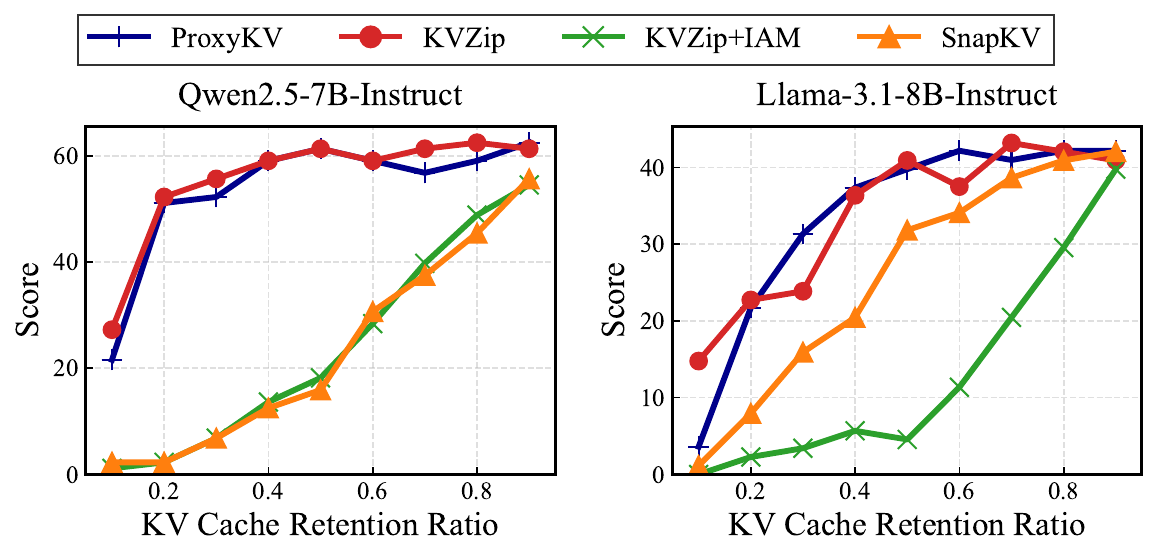}
        \caption{Zero-shot transfer to held-out SCBench \textit{RepoQA}. ProxyKV (blue) tracks the KVZip oracle (red) within $1$--$2$ pp on both targets. Left: Qwen-2.5; right: Llama-3.1.}
        \label{fig:scbench_repoqa_combined}
    \end{minipage}\hfill
    \begin{minipage}[c]{0.46\linewidth}
        \textbf{Robust zero-shot generalization.}
        \textit{ProxyKV transfers zero-shot to repository-level reasoning, matching KVZip on the held-out SCBench.RepoQA task.} As shown in \Cref{fig:scbench_repoqa_combined}, ProxyKV maintains parity with KVZip on both Llama-3.1 and Qwen-2.5 despite RepoQA being excluded from the training mixture (\textit{GSM8K}, \textit{SQuAD}, \textit{NIAH}, and a subset of SCBench). In the $\rho \in [0.2, 0.4]$ band, ProxyKV outperforms SnapKV and KVZip+IAM by a clear margin, evidencing the HybridAxialMapper's structural alignment.
    \end{minipage}
\end{figure}

\begin{figure}[t]
    \centering
    \includegraphics[width=\linewidth]{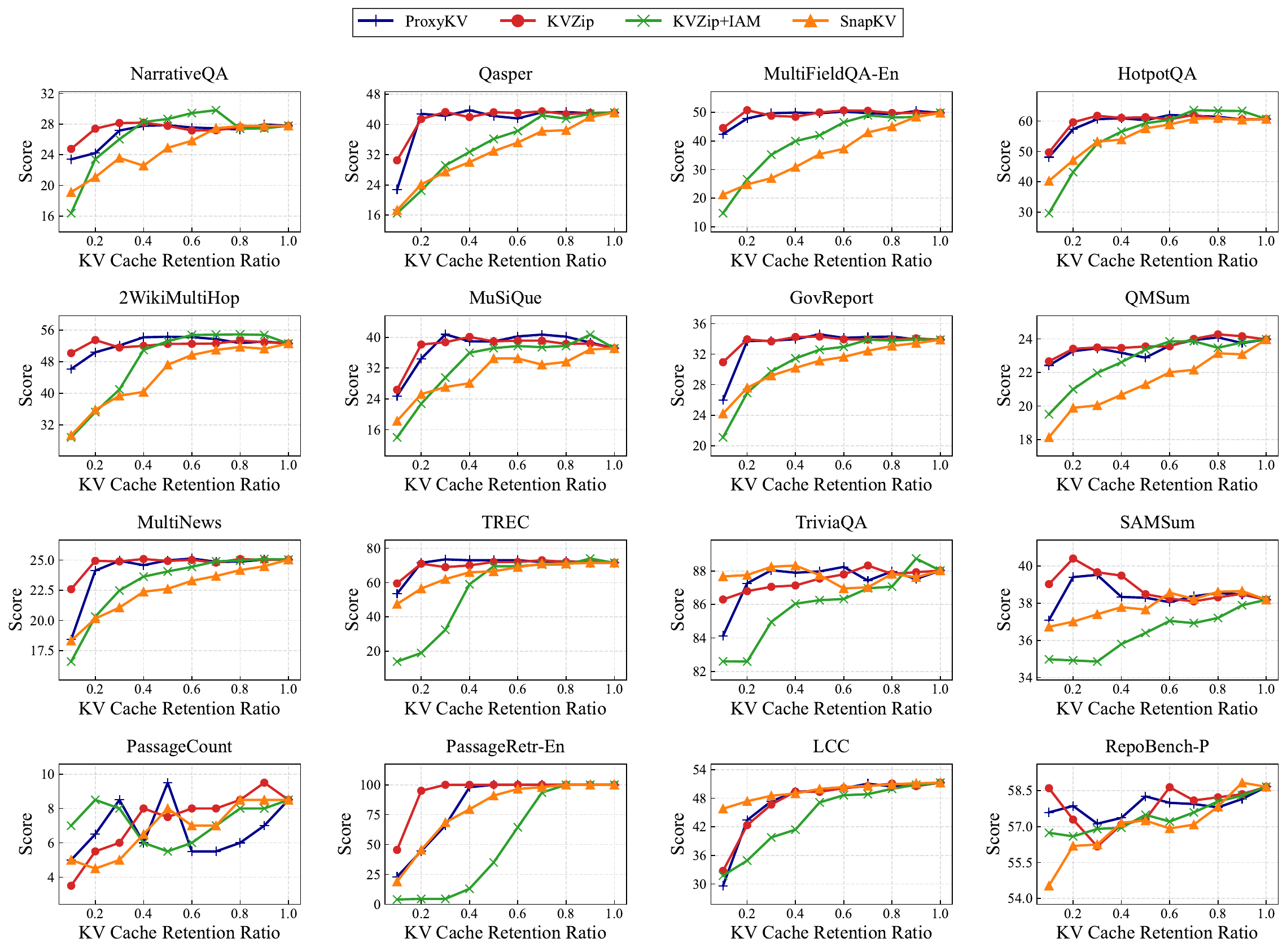}
    \caption{Per-dataset performance on the 16 English LongBench tasks (Qwen-2.5); the remaining 5 Chinese subsets are reported in \Cref{appendix:complete_longbench}. ProxyKV tracks the KVZip oracle and surpasses SnapKV on dense-synthesis tasks.}
    \label{fig:detailed_results}
\end{figure}

\textbf{Generalization to unseen datasets.}
\textit{ProxyKV matches or exceeds the KVZip oracle on the 16 English LongBench tasks despite training on a different distribution.} As shown in \Cref{fig:detailed_results}, ProxyKV (blue) tracks KVZip (red) on \textit{NarrativeQA} and \textit{MultiFieldQA-En} without the secondary prefilling penalty, and maintains high precision at aggressive pruning ratios on \textit{2WikiMultiHop} and \textit{QMSum} where SnapKV (orange) saturates. ProxyKV is near-lossless on code-intensive \textit{LCC} and \textit{RepoBench-P}; SnapKV remains competitive only on simple structured tasks like \textit{TREC} and fails on dense synthesis (\textit{SAMSum}). The full 21-subset breakdown including the 5 Chinese tasks is in \Cref{appendix:complete_longbench}.

\subsection{Efficiency analysis}

\begin{figure}[t]
    \centering
    \begin{subfigure}[t]{0.245\linewidth}\centering
        \includegraphics[width=\linewidth]{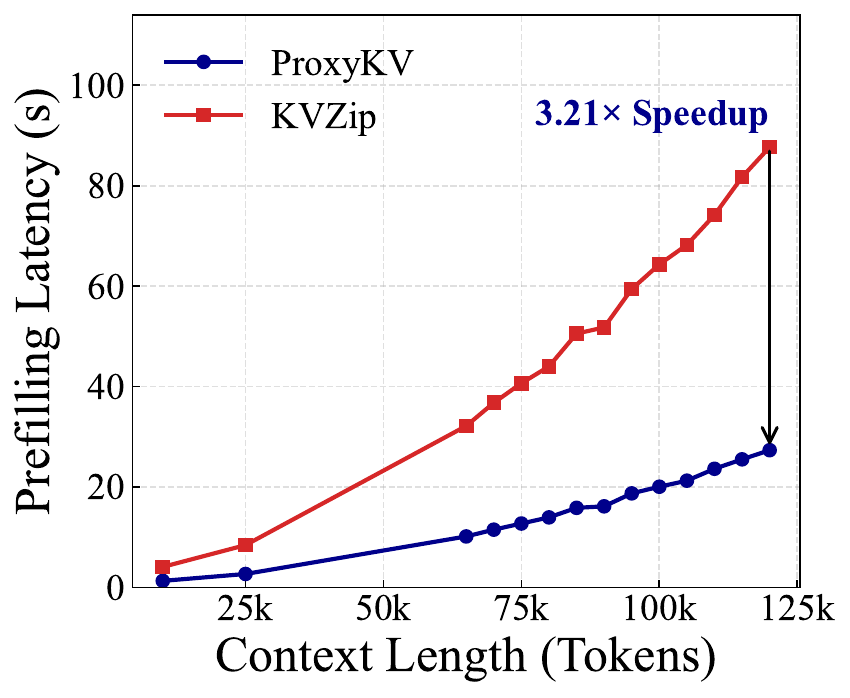}
        \caption{Latency, Llama-3.1-8B}
        \label{fig:latency_llama}
    \end{subfigure}\hfill
    \begin{subfigure}[t]{0.245\linewidth}\centering
        \includegraphics[width=\linewidth]{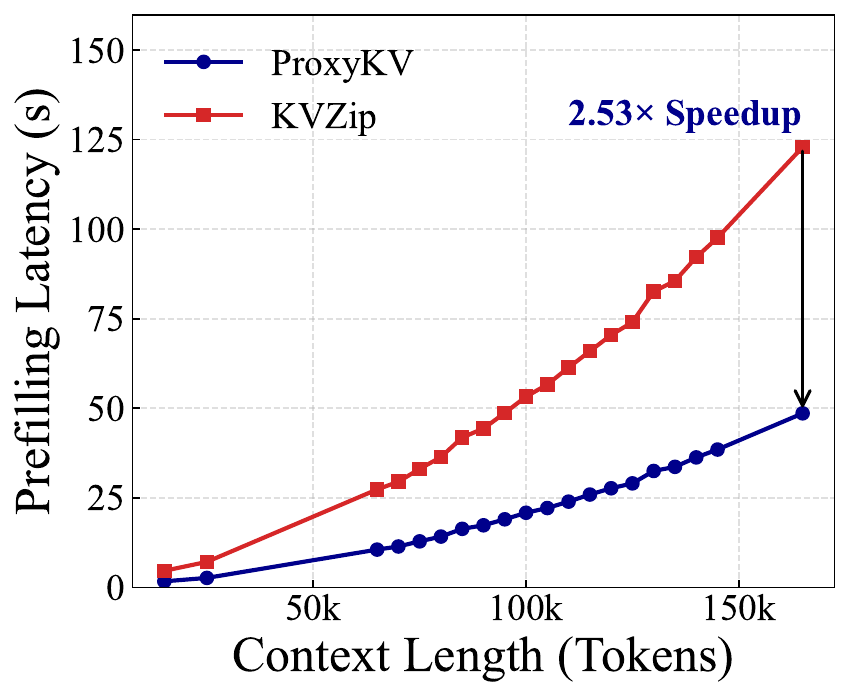}
        \caption{Latency, Qwen-2.5-7B}
        \label{fig:latency_qwen}
    \end{subfigure}\hfill
    \begin{subfigure}[t]{0.245\linewidth}\centering
        \includegraphics[width=\linewidth]{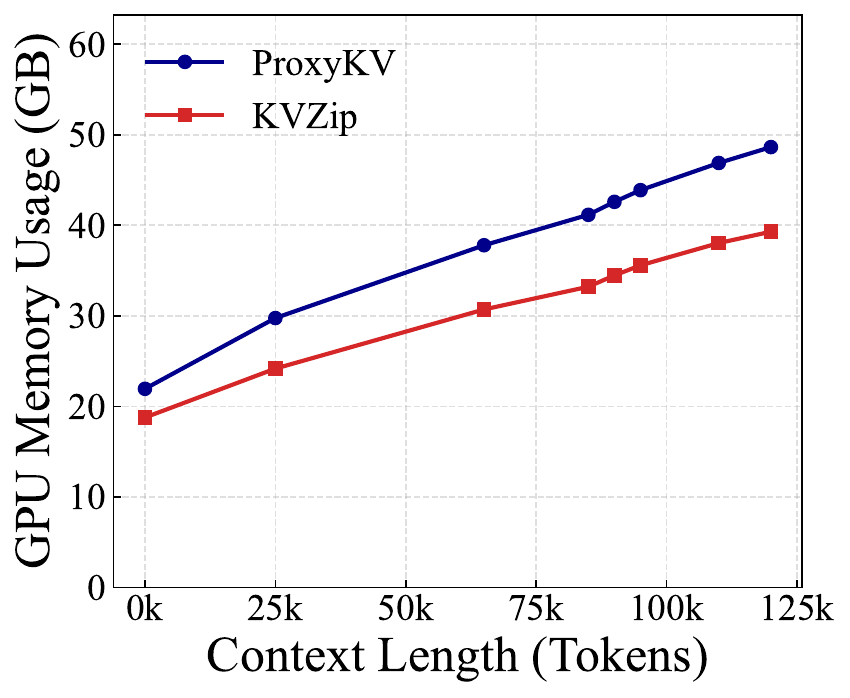}
        \caption{Memory, Llama-3.1-8B}
        \label{fig:memory_llama}
    \end{subfigure}\hfill
    \begin{subfigure}[t]{0.245\linewidth}\centering
        \includegraphics[width=\linewidth]{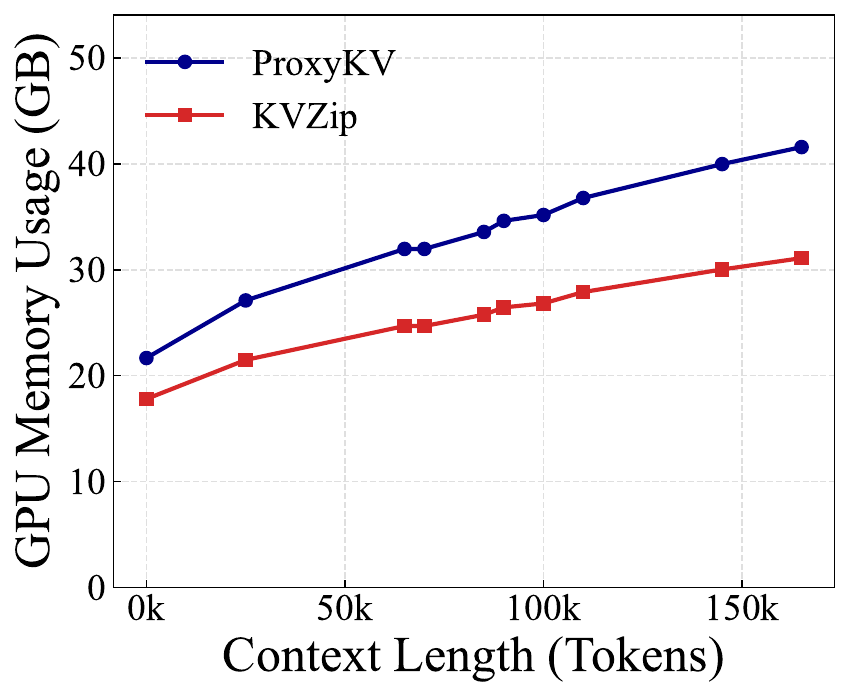}
        \caption{Memory, Qwen-2.5-7B}
        \label{fig:memory_qwen}
    \end{subfigure}
    \caption{ProxyKV flattens the super-linear latency curve of KVZip while paying a modest one-time memory premium. (a--b): prefilling latency; (c--d): peak GPU memory, across context length.}
    \label{fig:scalability_plot}
\end{figure}
\label{sec:efficiency}

\textbf{Latency scalability and memory trade-off.}
\textit{ProxyKV delivers up to $3.21\times$ prefilling speedup at a modest $16.91\%$--$33.72\%$ memory premium.} \Cref{fig:scalability_plot}(a--b) shows KVZip's secondary prefilling pass scales super-linearly up to 125k (Llama-3.1) and 170k (Qwen-2.5); ProxyKV flattens this curve by asynchronous offloading, yielding $3.21\times$ on Llama-3.1-8B and $2.53\times$ on Qwen-2.5-7B. The cost is the joint target+proxy footprint (\Cref{fig:scalability_plot}(c--d)); the proxy KV cache is freed at end of prefill, so the premium is a transient prefill-only peak (\Cref{fig:memory_timeline}, \Cref{appendix:memory_timeline}).

\begin{figure}[!t]
    \centering
    \includegraphics[width=0.92\linewidth]{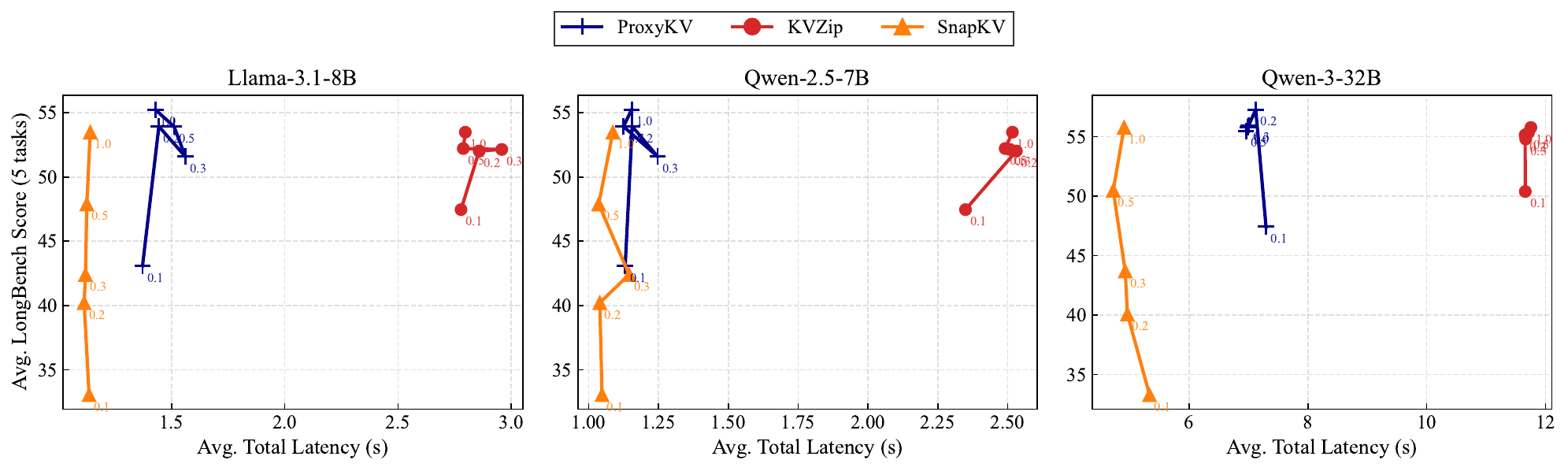}
    \caption{Score--latency Pareto on LongBench (5 representative tasks, total time = prefill + generation). Numbers next to each marker indicate retention ratio $\rho$. ProxyKV (blue) dominates KVZip (red) on latency at every $\rho$ and dominates SnapKV (orange) on score above $\sim$$1.3$\,s.}
    \label{fig:latency_pareto}
\end{figure}

\textbf{Score-latency Pareto on real LongBench.}
\textit{ProxyKV occupies the upper-left of the score--latency frontier on every model.} On Llama-3.1-8B at $\rho{=}0.5$, ProxyKV reaches $53.9$ at $1.51$\,s vs.\ KVZip's $52.2$ at $2.79$\,s ($1.85\times$ at near-equal accuracy); SnapKV collapses to $47.9$ at the same budget. On Qwen-3-32B, ProxyKV holds $55.8$ at $7.0$\,s vs.\ KVZip's $11.7$\,s ($1.68\times$ at equal score).

\begin{figure}[!t]
    \centering
    \includegraphics[width=0.92\linewidth]{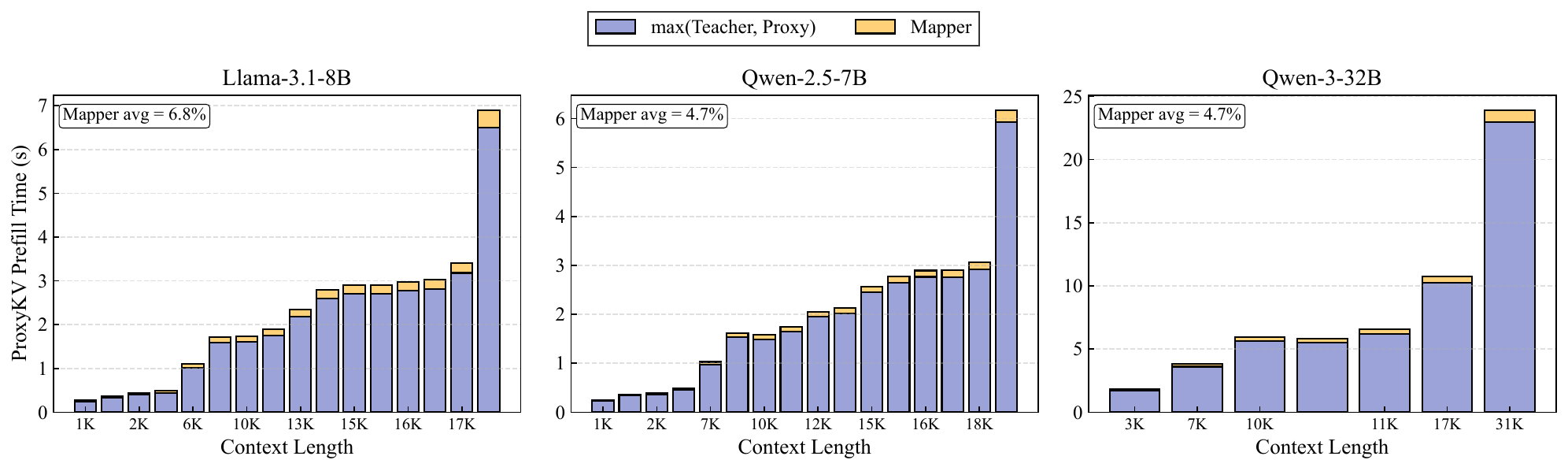}
    \caption{ProxyKV prefill breakdown at $\rho{=}0.3$. Teacher and proxy run in parallel; the visible bottom block is $\max(t_\text{teacher}, t_\text{proxy})$, while the HybridAxialMapper accounts for $4.7\%$--$6.8\%$ of wall time and remains a low-order term as the target scales from 7B (Qwen-2.5) and 8B (Llama-3.1) up to 32B (Qwen-3).}
    \label{fig:latency_mapper}
\end{figure}

\textbf{GPU-matched comparison and mapper overhead.}
\textit{ProxyKV's speedup is algorithmic, not hardware-parallel.} Co-locating proxy and teacher on a single GPU via CUDA streams still preserves $\sim$$1.5\times$ speedup over KVZip on Llama-3.1-8B and $\sim$$1.4\times$ on Qwen-2.5-7B across 1.6K--30K LongBench inputs (\Cref{appendix:realworld_latency}; 32B omitted as it must be sharded). The HybridAxialMapper itself contributes only $4.7\%$--$6.8\%$ of prefill wall time across 7B--32B targets (\Cref{fig:latency_mapper}), and the ratio cannot grow with target size because mapper compute scales with proxy width while the dominant teacher--proxy block scales with target width.

\setlength{\intextsep}{12pt plus 2pt minus 2pt}    
\setlength{\textfloatsep}{20pt plus 2pt minus 4pt}
\setlength{\floatsep}{12pt plus 2pt minus 2pt}
\makeatletter
\setlength{\@fpsep}{8pt plus 2fil minus 0pt}       
\setlength{\@fptop}{0pt plus 1fil}
\setlength{\@fpbot}{0pt plus 1fil}
\makeatother
\pagebreak
\subsection{Stress test on RULER}
\label{sec:ruler}

\begin{figure}[!t]
    \centering
    \includegraphics[width=0.82\linewidth]{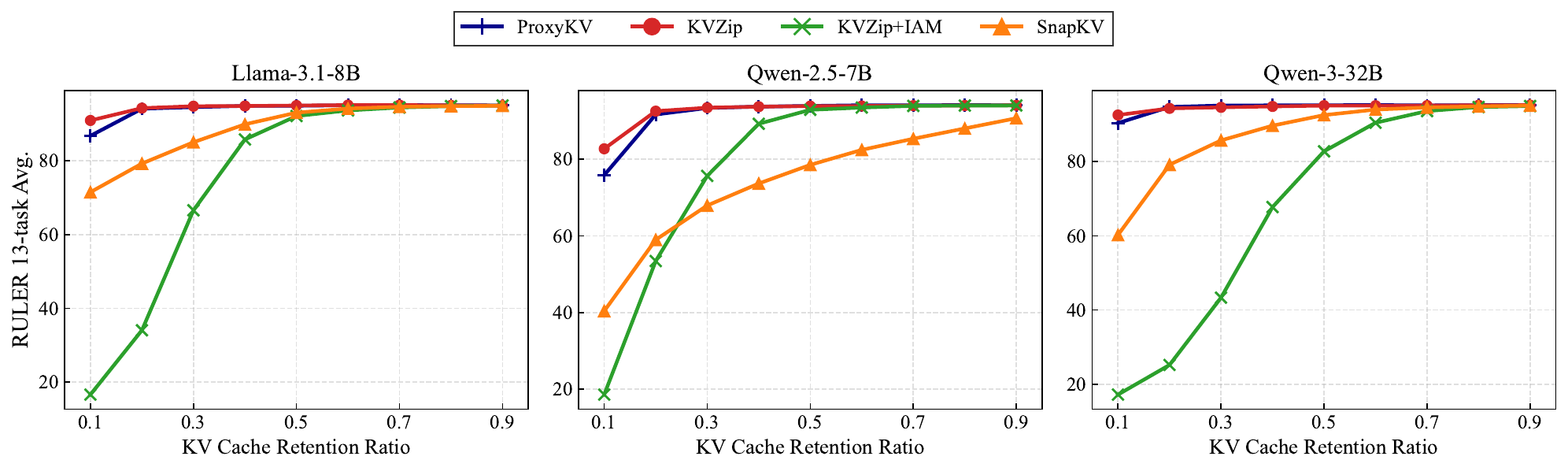}
    \caption{ProxyKV tracks the KVZip oracle on RULER across all three target scales (7B, 8B, 32B). RULER 13-task average score versus retention ratio $\rho$.}
    \label{fig:ruler_avg}
\end{figure}

\textit{ProxyKV preserves the KVZip-oracle RULER accuracy from 7B to 32B targets.} \Cref{fig:ruler_avg} averages the 13 RULER~\citep{hsieh2024ruler} subsets over $\rho \in [0.1, 0.9]$: ProxyKV tracks KVZip within $1$--$2$ points on every target, including the $\sim$$8\times$ Qwen-3-32B/Qwen-3-4B pair. SnapKV collapses on multi-needle/aggregation primitives at $\rho \leq 0.3$, while KVZip+IAM degrades \emph{most} on Qwen-3-32B (reaching $0$ on \emph{NIAH-MK3}/\emph{MV}/\emph{FWE} at $\rho{=}0.1$), confirming that rigid layer-to-layer alignment binds harder as the target scales. Per-task breakdowns: \Cref{appendix:ruler}.
\section{Ablation studies}
\label{sec:ablation}

We isolate the five loss terms, the three HybridAxialMapper stages, and the dominant loss coefficients. Since \Cref{fig:main_results,fig:ruler_avg} cluster tightly at $\rho \geq 0.5$, we focus on $\rho \in \{0.1, 0.2\}$. Ablations use Llama-3.1-8B / Llama-3.2-1B (each LOO variant retrained); the cross-family/cross-scale check is the Qwen results in \Cref{sec:ruler}. Component- and hyperparameter-LOOs and per-task swing tables are in \Cref{appendix:ablation_extra}; we summarise the loss-function LOO here.

\begin{wrapfigure}[11]{r}{0.45\linewidth}
    \vspace{-0.8\baselineskip}
    \centering
    \includegraphics[width=\linewidth]{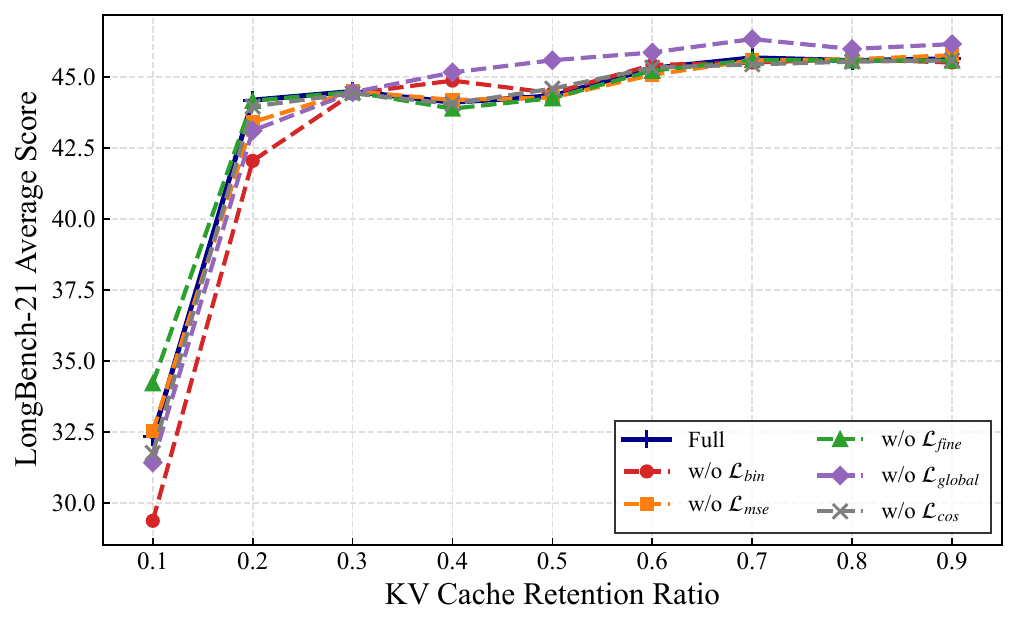}
    \captionsetup{font=footnotesize,skip=2pt}
    \caption{Loss LOO ablation, LongBench-21 average.}
    \label{fig:loss_loo_lines}
\end{wrapfigure}

\textit{$\mathcal{L}_{bin}$ is the single most critical term at low retention, and the five loss swing-supports are nearly disjoint.} At $\rho \in \{0.1, 0.2\}$ removing $\mathcal{L}_{bin}$ produces the largest drop; for $\rho \geq 0.4$ all six curves collapse into a $1$-point band. The per-task LOO on LongBench-21 (\Cref{tab:loss_task_owner}) shows nearly disjoint ownership at $\rho{=}0.1$: $\mathcal{L}_{bin}$ for sharp-cut (PassageRetr-En/Zh: $-11.0$/$-10.5$), $\mathcal{L}_{mse}$ for magnitude (TREC: $-15.5$), $\mathcal{L}_{cos}$ for anti-drift (DuReader: $-3.78$). The aggregate $1$-point band at $\rho \geq 0.4$ hides $25$-point per-task swings at $\rho \leq 0.2$; component and coefficient LOOs reach the same conclusion (\Cref{appendix:ablation_extra}).

\FloatBarrier
\section{Conclusion}

ProxyKV offloads importance scoring to an asynchronous Small-Model Proxy via the HybridAxialMapper and a ranking-consistent hybrid loss, matching KVZip across Llama-3.1, Qwen-2.5, and Qwen-3 ($\sim$$98.7\%$ mean recovery, $7$B--$32$B targets) at up to $3.21\times$ prefilling speedup (dual-GPU; $\sim$$1.5\times$ shared single-GPU), sustained to $170$k tokens; the ProxyKV--IAM gap \emph{widens} with target scale, exposing rigid layer-to-layer alignment as the binding constraint that the learnable cross-axial mapping removes.

\paragraph{Limitations.}
ProxyKV requires intra-family pairs and raises peak-prefill GPU memory by $16.91\%$--$33.72\%$ (transient proxy activations released after prefill, \Cref{appendix:memory_timeline}); in vLLM/TGI continuous batching this peak still caps concurrent batch size. Cross-family transfer is future work.

\paragraph{Broader Impact.}
ProxyKV lowers long-context inference cost; potential misuse on sensitive corpora warrants safety monitoring and access controls.

\bibliographystyle{plainnat}
\bibliography{references}

\clearpage
\appendix

\section{Real-world LongBench latency analysis}
\label{appendix:realworld_latency}

This appendix expands the score--latency Pareto plot of \Cref{fig:latency_pareto} (\Cref{sec:efficiency}) with two robustness checks: a single-GPU context-length scan that constrains every method to the same hardware budget, and a per-stage prefill breakdown isolating the HybridAxialMapper's overhead. SnapKV is included as the heuristic floor; KVZip+IAM is omitted because its in-place layer alignment offers no measurable runtime difference from KVZip on the prefill path.

\begin{figure}[!ht]
    \centering
    \includegraphics[width=\linewidth]{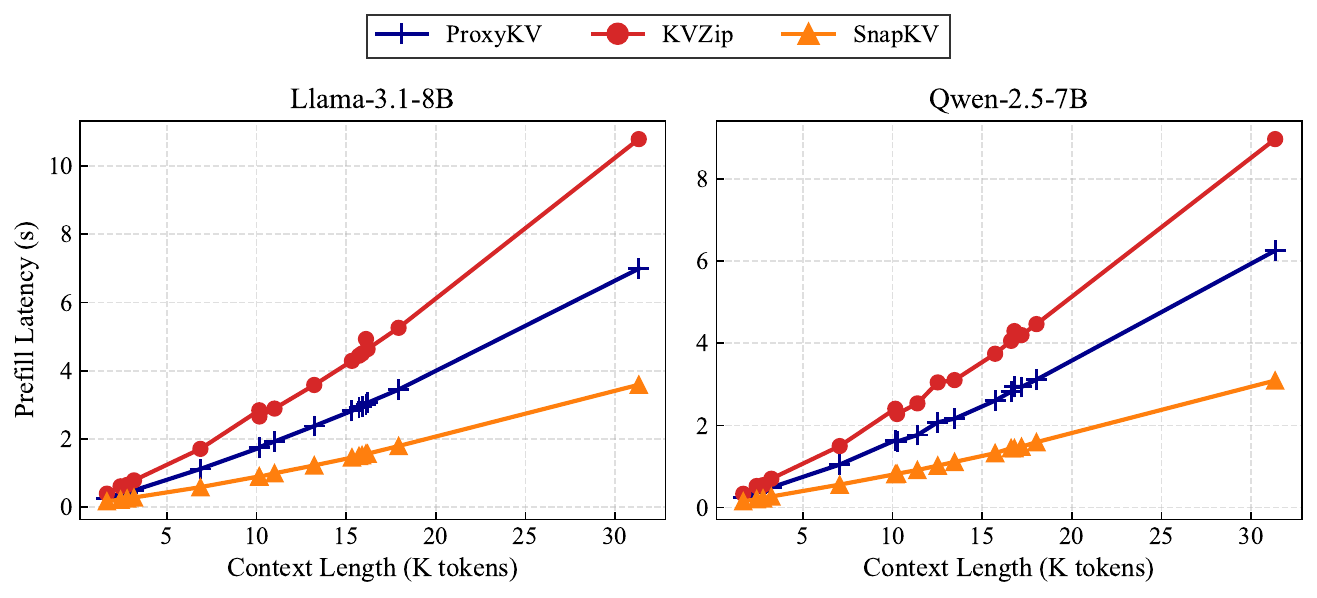}
    \caption{Single-GPU prefill latency vs.\ context length on real LongBench inputs at $\rho{=}0.3$ for Llama-3.1-8B and Qwen-2.5-7B. The ProxyKV--KVZip gap widens monotonically with context length.}
    \label{fig:latency_ctxlen}
\end{figure}

\paragraph{GPU-count-matched single-GPU context scan.}
\textit{ProxyKV remains $1.3\times$--$1.6\times$ faster than KVZip when every method is constrained to the same single GPU.} \Cref{fig:latency_ctxlen} controls for device count on the two targets that fit on a single GPU by running ProxyKV's teacher and proxy concurrently via CUDA streams on the \emph{same} GPU as KVZip and SnapKV, isolating the algorithmic speedup from hardware parallelism. Across 22 LongBench samples per model spanning $1.6$K to $30$K tokens---the upper bound at which all four methods fit on a single RTX PRO 6000 without OOM---ProxyKV preserves a $\sim$$1.5\times$ speedup over KVZip on Llama-3.1-8B and a $\sim$$1.4\times$ speedup on Qwen-2.5-7B across the entire context-length range. The matched-device comparison is omitted for Qwen-3-32B because the 32B target must be sharded across multiple GPUs, making ``1 GPU per method'' undefined. The per-stage prefill breakdown that quantifies the mapper's $4.7\%$--$6.8\%$ wall-time share is reproduced in \Cref{fig:latency_mapper} of the main text.

\section{Memory timeline of the dual-GPU pipeline}
\label{appendix:memory_timeline}

This appendix complements the per-context memory footprint of \Cref{fig:scalability_plot}(c--d) (\Cref{sec:efficiency}) with a per-GPU \emph{memory--time} trace that locates the $16.91\%$--$33.72\%$ memory premium in time. The aggregate plots in the main text report the joint footprint at a single (worst-case) instant; the trace below shows that this instant is confined to the prefill window and that the proxy-side overhead vanishes during decode. We sample per-GPU usage every $50$\,ms via NVML while running the Llama-3.1-8B target on GPU\,1 and the Llama-3.2-1B proxy on GPU\,2 over an $8$K-token context, followed by $32$ decode steps. The trace separates the pipeline into three phases: (i)~weight loading, (ii)~parallel prefill (target and proxy concurrent), and (iii)~decode on the target only.

\begin{figure}[!ht]
    \centering
    \includegraphics[width=0.92\linewidth]{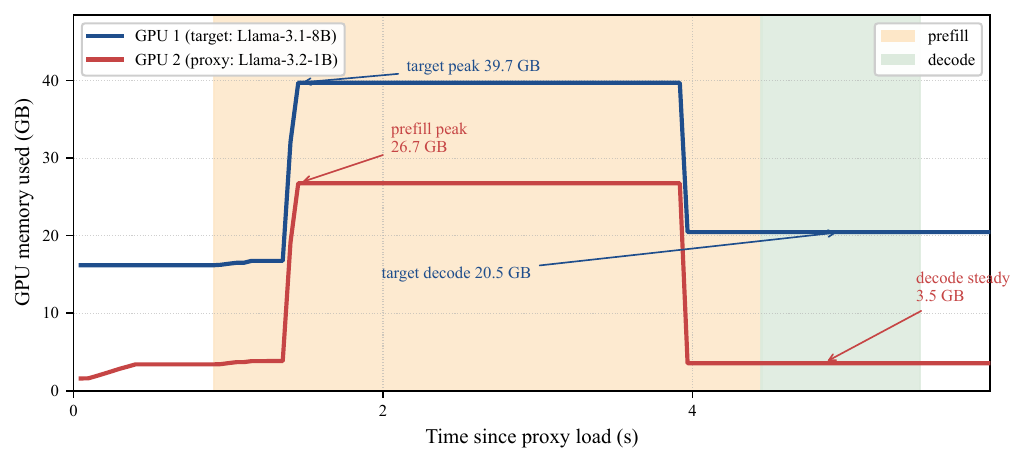}
    \caption{Dual-GPU memory timeline for ProxyKV (Llama-3.1-8B target on GPU\,1, Llama-3.2-1B proxy on GPU\,2). The orange band marks the prefill phase, the green band marks decode. The proxy GPU jumps from $3.5$\,GB (weights only) to $26.7$\,GB at the prefill peak---driven almost entirely by the prefill-time activation working set (attention logits, intermediate projections, and short-lived hidden states), since the $1$B proxy's KV cache itself is well under $1$\,GB at this context length---then drops back to $3.5$\,GB the instant the prefill activations and the proxy KV cache are released, and stays flat for the entire decode window. The $\sim$$23$\,GB premium is therefore a transient prefill-only peak, not a sustained cost.}
    \label{fig:memory_timeline}
\end{figure}

\paragraph{Prefill-only transient peak.}
\textit{The proxy-side memory overhead is bounded to the prefill window and is released before decode starts.} As shown in \Cref{fig:memory_timeline}, GPU\,2 (proxy) climbs from $3.5$\,GB (weights only) to $26.7$\,GB during prefill. The dominant contributor to this $\sim$$23$\,GB spike is \emph{not} the proxy KV cache (which, for a $1$B-parameter model in $16$-bit precision, occupies under $1$\,GB even at this context length) but the proxy's prefill activation working set---attention logit matrices, intermediate Q/K/V/FFN projections, and the auxiliary tensors required by the HybridAxialMapper to extract the cross-head feature $\mathbf{X}$. The instant prefill finishes, both the proxy KV cache and the activation working set are released, and GPU\,2 returns to $3.5$\,GB for the entire decode phase, a $\sim$$7.6\times$ reduction. GPU\,1 (target) follows a parallel pattern: it peaks at $39.7$\,GB during prefill (target weights $+$ target prefill activations $+$ full target KV cache) and settles at $20.5$\,GB during decode after the pruning mask is applied. To make this distinction explicit, the $16.91\%$--$33.72\%$ memory premium reported in \Cref{sec:efficiency} is the \emph{peak prefill} GPU footprint relative to KVZip, including activations; the corresponding theoretical-only premium (weights $+$ retained KV cache) is much smaller. We note that this peak is the load-bearing quantity for serving deployments: in continuous-batching engines such as vLLM and TGI, where prefill and decode share a common GPU memory pool, the prefill peak must be reservable for any concurrent request, so the maximum batch size is set by the peak rather than by the time-averaged footprint.

\section{Detailed ablation analysis}
\label{appendix:ablation_extra}

\begin{figure}[!ht]
    \centering
    \includegraphics[width=0.7\linewidth]{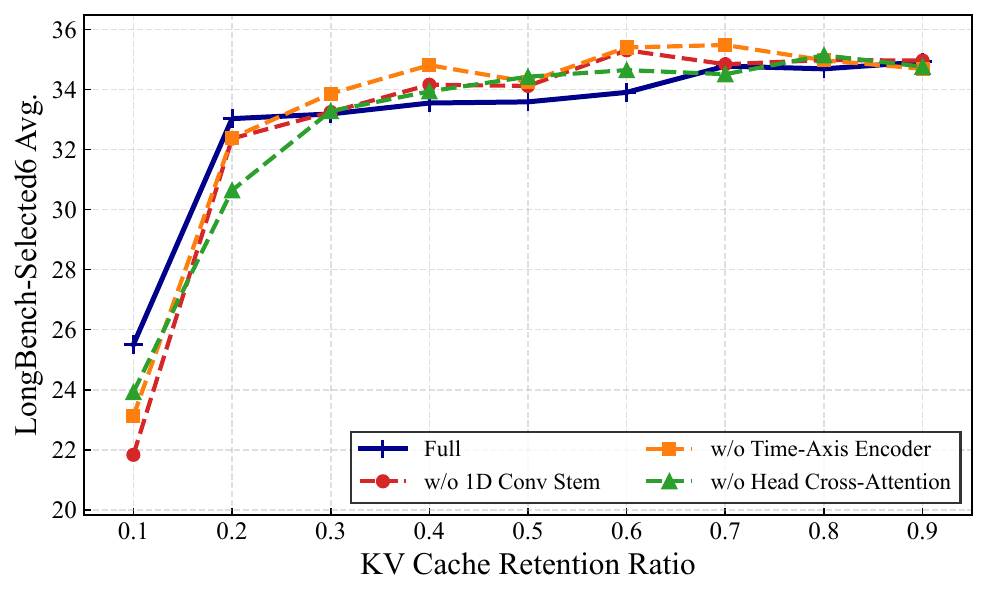}
    \caption{Component leave-one-out on the six representative LongBench subsets, six-task average vs.\ retention ratio $\rho$. The discriminating regime is $\rho \in \{0.1, 0.2\}$, where ``w/o Conv'' incurs the largest drop ($-3.67$ on the six-task average), followed by ``w/o Time'' ($-2.38$) and ``w/o Head'' ($-1.58$); all four configurations re-converge once $\rho \geq 0.5$. Per-task drops are quantified in \Cref{tab:abl_components_per_task}.}
    \label{fig:abl_components}
\end{figure}

\begin{figure}[!ht]
    \centering
    \includegraphics[width=0.92\linewidth]{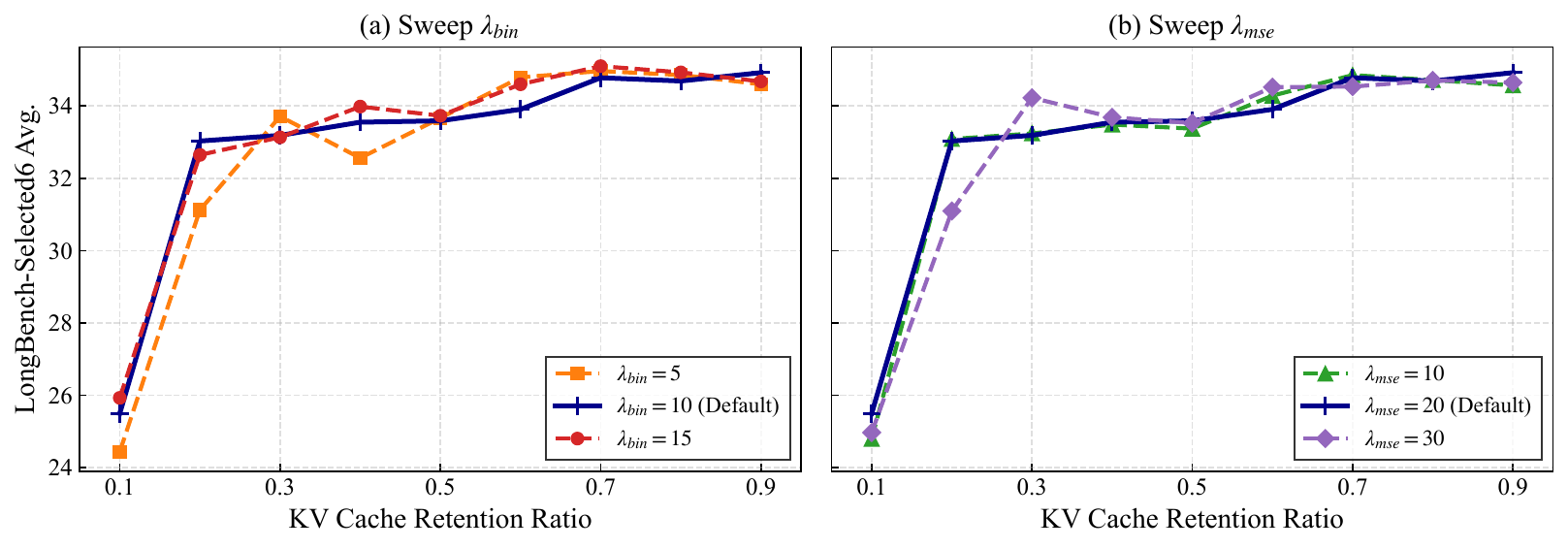}
    \caption{Hyperparameter sensitivity of the two dominant loss coefficients on the six representative LongBench subsets. The default $(\lambda_{bin}, \lambda_{mse}) = (10, 20)$ leads at $\rho \in \{0.1, 0.2\}$, and all sweeps reconverge at high retention.}
    \label{fig:abl_hyper}
\end{figure}

This appendix complements the loss-LOO summary in \Cref{sec:ablation} with (i) the per-task LOO swing tables across LongBench-21 (\Cref{tab:loss_task_map,tab:loss_task_owner}), (ii) the component LOO of the three HybridAxialMapper stages (\Cref{fig:abl_components}, \Cref{tab:abl_components_per_task}), and (iii) the hyperparameter sensitivity sweep over $\lambda_{bin}, \lambda_{mse}$ (\Cref{fig:abl_hyper}).

\begin{table}[!ht]
    \centering
    \footnotesize
    \setlength{\tabcolsep}{4pt}
    \caption{\textbf{Qualitative overview} of each loss term's intended role and the task families we hypothesised it would dominate prior to running the leave-one-out study. The ``Dominant tasks'' column lists representative datasets where the term is expected to matter; the actual per-task LOO winners at the diagnostic retention ratios are reported in \Cref{tab:loss_task_owner} and should be treated as the quantitative ground truth (a few tasks change owner between the two tables, which is informative about how the loss landscape shifts with $\rho$ rather than a contradiction in our claims). ``Diag.\ $\rho$'' marks the retention range where the term measurably moves the score; entries outside that are within noise. Dataset abbreviations: NQA = NarrativeQA, TQA = TriviaQA, HQA = HotpotQA, MFQA-Zh = MultiField-QA-Zh, PR-En/Zh = PassageRetr-En/Zh, QMS = QMSum, GR = GovReport, SAM = SAMSum, MN = MultiNews, PC = PassageCount, MuS = MuSiQue, DR = DuReader.}
    \label{tab:loss_task_map}
    \begin{tabular*}{\linewidth}{@{\extracolsep{\fill}}l l l l c@{}}
        \toprule
        Loss & Role & Dominant tasks & Why it matters & Diag.\ $\rho$ \\
        \midrule
        $\mathcal{L}_{bin}$    & Binary vital-token classifier        & TREC, PR-En, NQA   & Sharp-cut, few vitals        & $0.1$--$0.2$ \\
        $\mathcal{L}_{mse}$    & Value-weighted magnitude regression  & TQA, MFQA-Zh, HQA  & Magnitudes drive aggregation & $0.1$ \\
        $\mathcal{L}_{fine}$   & Intra-Top-$K$ pairwise rank          & QMS, GR, SAM, MN   & Order among kept tokens      & $0.2$--$0.5$ \\
        $\mathcal{L}_{global}$ & Mass-aware Top-$K$ separation        & PC, MuS            & Signal/noise boundary        & $0.1$ \\
        $\mathcal{L}_{cos}$    & Directional cosine alignment         & PR-Zh, DR, LSHT    & Anti-drift regularizer       & $0.1$--$0.3$ \\
        \bottomrule
    \end{tabular*}
\end{table}

\begin{table}[!ht]
    \centering
    \footnotesize
    \setlength{\tabcolsep}{4pt}
    \renewcommand{\arraystretch}{1.05}
    \caption{\textbf{Per-task most-needed loss term} on the LongBench-21 suite. For every task we report the LOO variant whose removal causes the largest score drop versus \emph{Full} at the diagnostic retention ratios $\rho{=}0.1$ and $\rho{=}0.2$, with the magnitude of that drop ($\Delta$, points). Drops with magnitude $\geq 3$ pts are bolded. Tasks are grouped by which loss term owns them at $\rho{=}0.1$, with a final \emph{non-diagnostic} block listing tasks whose five LOO variants all stay within $\pm 0.5$ pt of \emph{Full} at $\rho{=}0.1$. Abbreviations follow \Cref{tab:loss_task_map}.}
    \label{tab:loss_task_owner}
    \begin{tabular*}{\linewidth}{@{\extracolsep{\fill}}l l r l r@{}}
        \toprule
        Task & Top loss @ $\rho{=}0.1$ & $\Delta_{0.1}$ & Top loss @ $\rho{=}0.2$ & $\Delta_{0.2}$ \\
        \midrule
        \multicolumn{5}{@{}l}{\emph{$\mathcal{L}_{bin}$-dominated (sharp-cut / single-needle / multi-hop chains)}} \\
        PassageRetr-En     & $\mathcal{L}_{bin}$    & $\mathbf{-11.00}$ & $\mathcal{L}_{bin}$    & $-2.50$ \\
        PassageRetr-Zh     & $\mathcal{L}_{bin}$    & $\mathbf{-10.50}$ & $\mathcal{L}_{bin}$    & $\mathbf{-3.00}$ \\
        MultiField-QA-Zh   & $\mathcal{L}_{bin}$    & $\mathbf{-7.05}$  & $\mathcal{L}_{global}$ & $-1.52$ \\
        LSHT               & $\mathcal{L}_{bin}$    & $\mathbf{-5.25}$  & $\mathcal{L}_{mse}$    & $\mathbf{-6.00}$ \\
        2WikiMultiHop      & $\mathcal{L}_{bin}$    & $\mathbf{-4.88}$  & $\mathcal{L}_{bin}$    & $\mathbf{-3.67}$ \\
        MuSiQue            & $\mathcal{L}_{bin}$    & $\mathbf{-3.62}$  & $\mathcal{L}_{mse}$    & $-1.33$ \\
        HotpotQA           & $\mathcal{L}_{bin}$    & $-2.53$           & $\mathcal{L}_{fine}$   & $-0.15$ \\
        LCC (code)         & $\mathcal{L}_{bin}$    & $-2.39$           & $\mathcal{L}_{bin}$    & $-2.21$ \\
        GovReport          & $\mathcal{L}_{bin}$    & $-2.13$           & $\mathcal{L}_{global}$ & $-0.39$ \\
        MultiNews          & $\mathcal{L}_{bin}$    & $-1.32$           & $\mathcal{L}_{global}$ & $-0.26$ \\
        \midrule
        \multicolumn{5}{@{}l}{\emph{$\mathcal{L}_{mse}$-dominated (magnitudes drive aggregation)}} \\
        TREC               & $\mathcal{L}_{mse}$    & $\mathbf{-15.50}$ & $\mathcal{L}_{bin}$    & $\mathbf{-25.50}$ \\
        MultiField-QA-En   & $\mathcal{L}_{mse}$    & $-1.15$           & $\mathcal{L}_{global}$ & $\mathbf{-3.48}$ \\
        \midrule
        \multicolumn{5}{@{}l}{\emph{$\mathcal{L}_{global}$-dominated (Top-$K$ / non-Top-$K$ boundary)}} \\
        Qasper             & $\mathcal{L}_{global}$ & $-1.38$           & $\mathcal{L}_{bin}$    & $\mathbf{-3.68}$ \\
        NarrativeQA        & $\mathcal{L}_{global}$ & $-0.55$           & $\mathcal{L}_{mse}$    & $-1.12$ \\
        \midrule
        \multicolumn{5}{@{}l}{\emph{$\mathcal{L}_{fine}$-dominated (intra-Top-$K$ order; code completion)}} \\
        RepoBench-P        & $\mathcal{L}_{fine}$   & $-0.67$           & $\mathcal{L}_{global}$ & $-1.30$ \\
        \midrule
        \multicolumn{5}{@{}l}{\emph{$\mathcal{L}_{cos}$-dominated (anti-drift; long-tail Chinese / multilingual)}} \\
        DuReader           & $\mathcal{L}_{cos}$    & $\mathbf{-3.78}$  & $\mathcal{L}_{bin}$    & $-2.90$ \\
        TriviaQA           & $\mathcal{L}_{cos}$    & $-1.51$           & $\mathcal{L}_{global}$ & $-1.33$ \\
        SAMSum             & $\mathcal{L}_{cos}$    & $-0.55$           & $\mathcal{L}_{global}$ & $-0.82$ \\
        \midrule
        \multicolumn{5}{@{}l}{\emph{Non-diagnostic at $\rho{=}0.1$ (all five LOO variants within $\pm 0.5$ pt of \emph{Full})}} \\
        QMSum              & ---                    & ---               & $\mathcal{L}_{bin}$    & $-0.80$ \\
        PassageCount       & ---                    & ---               & $\mathcal{L}_{bin}$    & $-2.34$ \\
        VCSUM              & ---                    & ---               & $\mathcal{L}_{bin}$    & $-0.40$ \\
        \bottomrule
    \end{tabular*}
\end{table}

\paragraph{Component ablation.}
The HybridAxialMapper is a three-stage pipeline: (i) 1D Conv Stem, (ii) Time-Axis Encoder, (iii) Head-Axis Cross-Attention. We disable one stage at a time, replacing it with an identity (Conv Stem, Cross-Attention) or a mean-pool (Time-Axis Encoder), keeping all other parameters unchanged. The simplifying replacements preserve tensor shapes so that the loss objective is unmodified.

\begin{table}[!ht]
\centering
\caption{Component LOO at the discriminating regime $\rho \in \{0.1, 0.2\}$ on the six representative LongBench subsets. \emph{Full} columns are absolute ProxyKV accuracy; \emph{w/o} columns give the change ($\Delta$) when the corresponding HybridAxialMapper stage is disabled and re-trained. Drops with magnitude $\geq 3$ pts are bolded.}
\label{tab:abl_components_per_task}
\small
\setlength{\tabcolsep}{4pt}
\begin{tabular*}{\linewidth}{@{\extracolsep{\fill}}lcrrrcrrr@{}}
\toprule
& \multicolumn{4}{c}{$\rho{=}0.1$} & \multicolumn{4}{c}{$\rho{=}0.2$} \\
\cmidrule(lr){2-5}\cmidrule(lr){6-9}
Task & Full & w/o Conv & w/o Time & w/o Head & Full & w/o Conv & w/o Time & w/o Head \\
\midrule
2WikiMQA      & 34.69 & \textbf{$-4.70$} & $-1.32$ & $-2.23$ & 44.36 & $-0.25$ & $+1.54$ & $-0.41$ \\
NarrativeQA   & 23.65 & $+0.16$ & $+1.40$ & $+0.42$ & 29.30 & $-1.09$ & $-0.77$ & $-0.50$ \\
PassageCount  &  0.78 & $+1.22$ & $-0.78$ & $+0.22$ &  3.34 & $+2.19$ & $+0.33$ & $-1.19$ \\
QMSum         & 22.02 & $-0.22$ & $+0.93$ & $+1.01$ & 24.83 & $-0.99$ & $-0.66$ & $-0.05$ \\
RepoBench-P   & 43.87 & $+1.01$ & $+0.98$ & $-0.92$ & 43.83 & $-0.83$ & $+0.68$ & $-1.18$ \\
TREC          & 28.00 & \textbf{$-19.50$} & \textbf{$-15.50$} & \textbf{$-8.00$} & 52.50 & \textbf{$-3.00$} & \textbf{$-5.00$} & \textbf{$-11.00$} \\
\midrule
\textbf{6-Avg} & \textbf{25.50} & \textbf{$-3.67$} & $-2.38$ & $-1.58$ & \textbf{33.03} & $-0.66$ & $-0.65$ & $-2.39$ \\
\bottomrule
\end{tabular*}
\end{table}

\textit{The full design unlocks model usability at $\rho{=}0.2$, where every stage is load-bearing.} The complete HybridAxialMapper reaches $33.03$ on the selected-six average at $\rho{=}0.2$, recovering $94.5\%$ of the uncompressed reference ($34.94$ at $\rho{=}1.0$) while discarding $80\%$ of the KV cache. The per-task evidence in \Cref{tab:abl_components_per_task} concentrates the aggregate drop on TREC ($-19.5$/$-15.5$/$-8.0$ at $\rho{=}0.1$ for w/o Conv/Time/Head) and 2WikiMQA ($-4.7$ w/o Conv); the three-stage architecture is most necessary on tasks that require sharp-cut classification of vital tokens or multi-hop alignment at aggressive compression.

\paragraph{Hyperparameter sensitivity.}
We sweep the two dominant loss coefficients $\lambda_{bin}$ and $\lambda_{mse}$ around their default values ($\lambda_{bin}{=}10$, $\lambda_{mse}{=}20$). For each sweep we hold all other coefficients fixed and retrain. The remaining coefficients ($\lambda_{fine}{=}3$, $\lambda_{global}{=}2$, $\lambda_{cos}{=}0.5$) are an order of magnitude smaller and were not found to be performance-critical in pilot studies.

\textit{The default coefficients lead at $\rho \in \{0.1, 0.2\}$ and the sweep is uninformative above $\rho{=}0.4$.} In the high-retention tail $\rho \in [0.4, 0.9]$ the curves overlap within $0.6$ average points and cannot rank the candidates. At $\rho \in \{0.1, 0.2\}$ the spread widens (up to $1.5$ points), with a $\sim$$2\times$ perturbation in either direction degrading by less than $1$ point. We attribute the residual robustness to the multi-ratio nature of $\mathcal{L}_{bin}$: even at $\lambda_{bin}{=}5$ the binary signal still receives roughly a quarter of the gradient budget.

\section{Complete LongBench results}
\label{appendix:complete_longbench}

\begin{figure}[!htbp]
    \centering
    \includegraphics[width=\linewidth]{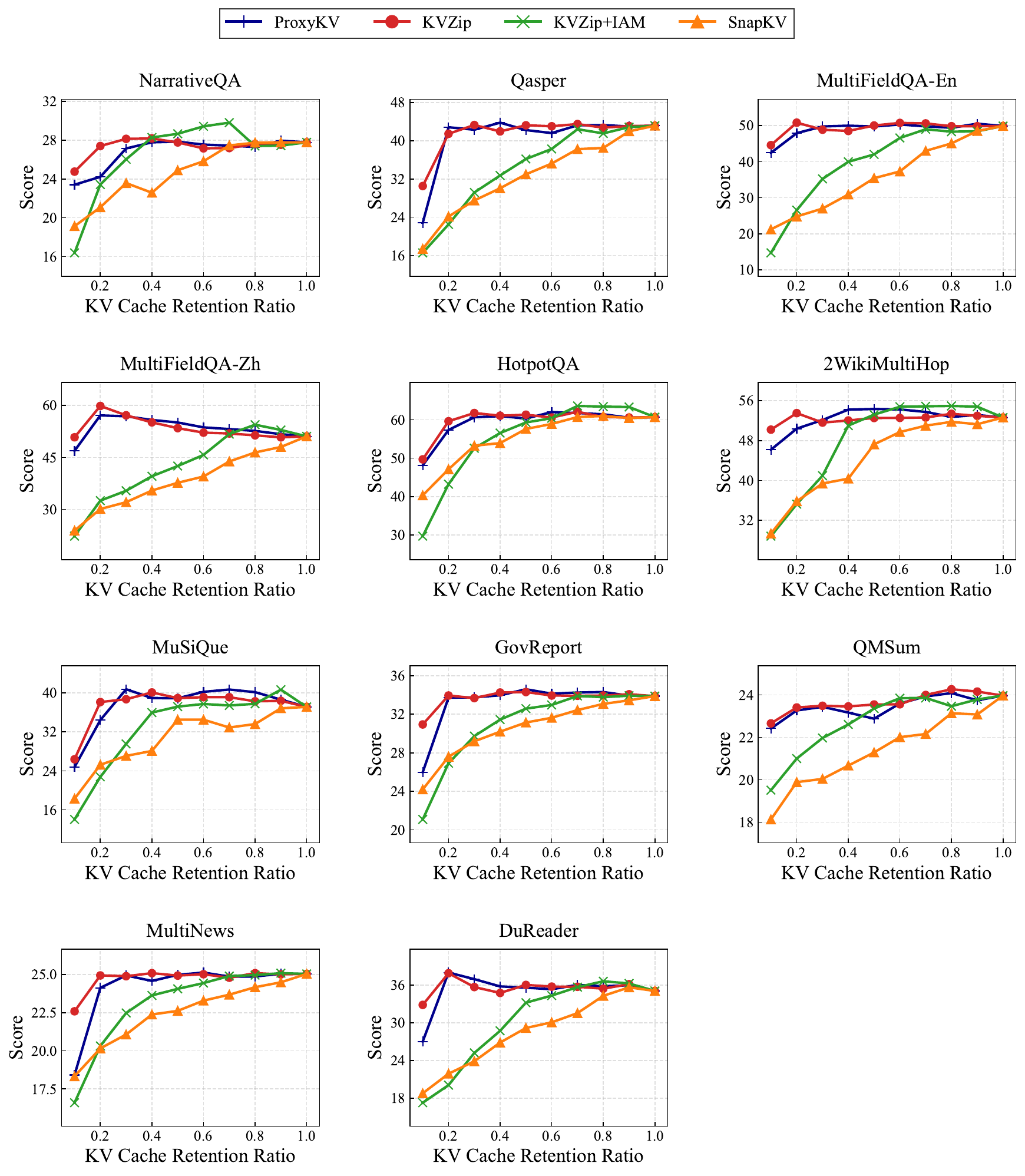}
    \caption{Complete LongBench results, Qwen-2.5, Part I.}
    \label{fig:longbench_21_qwen_part1}
\end{figure}

\begin{figure}[!htbp]
    \centering
    \includegraphics[width=\linewidth]{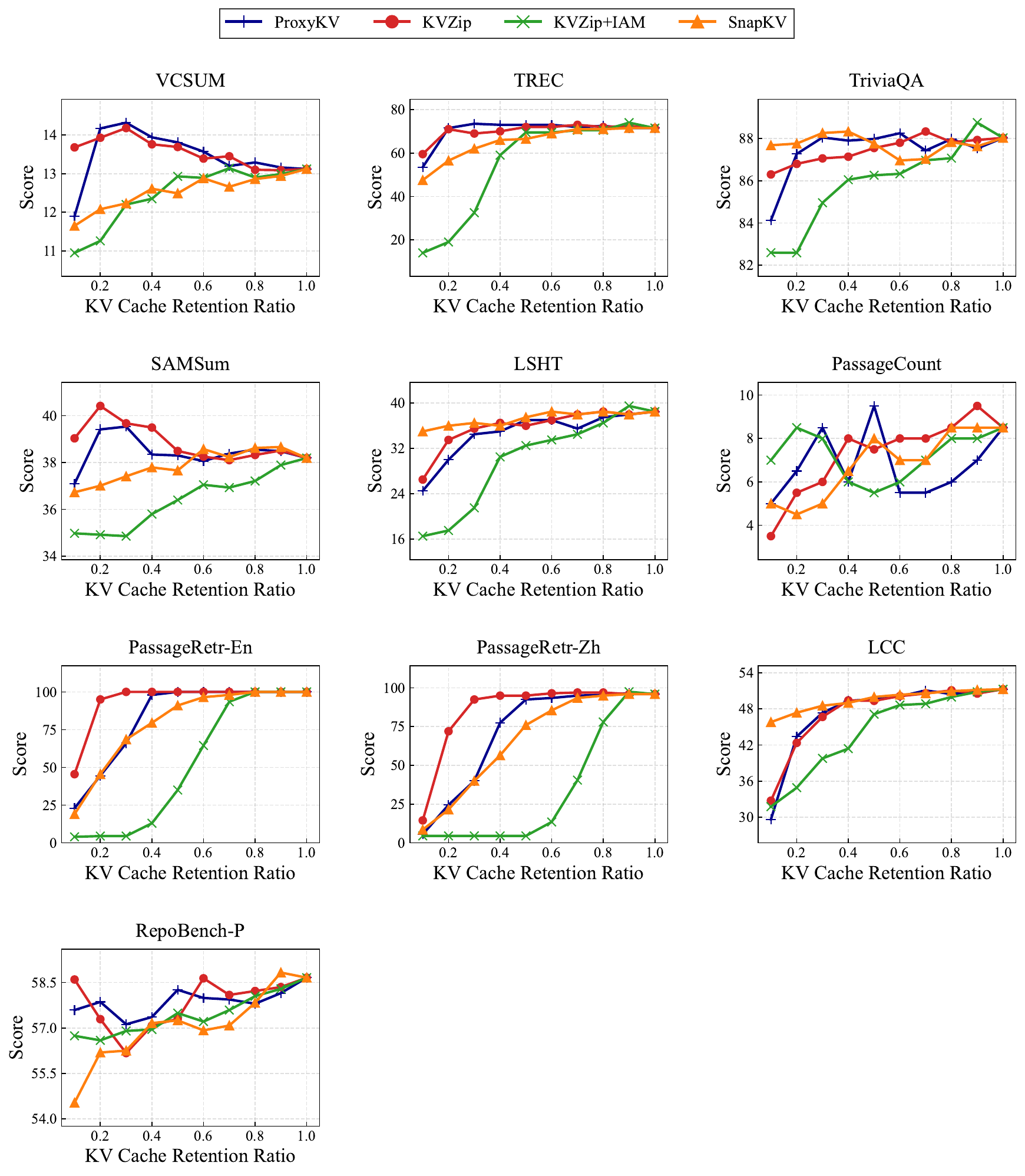}
    \caption{Complete LongBench results, Qwen-2.5, Part II.}
    \label{fig:longbench_21_qwen_part2}
\end{figure}

\begin{figure}[!htbp]
    \centering
    \includegraphics[width=\linewidth]{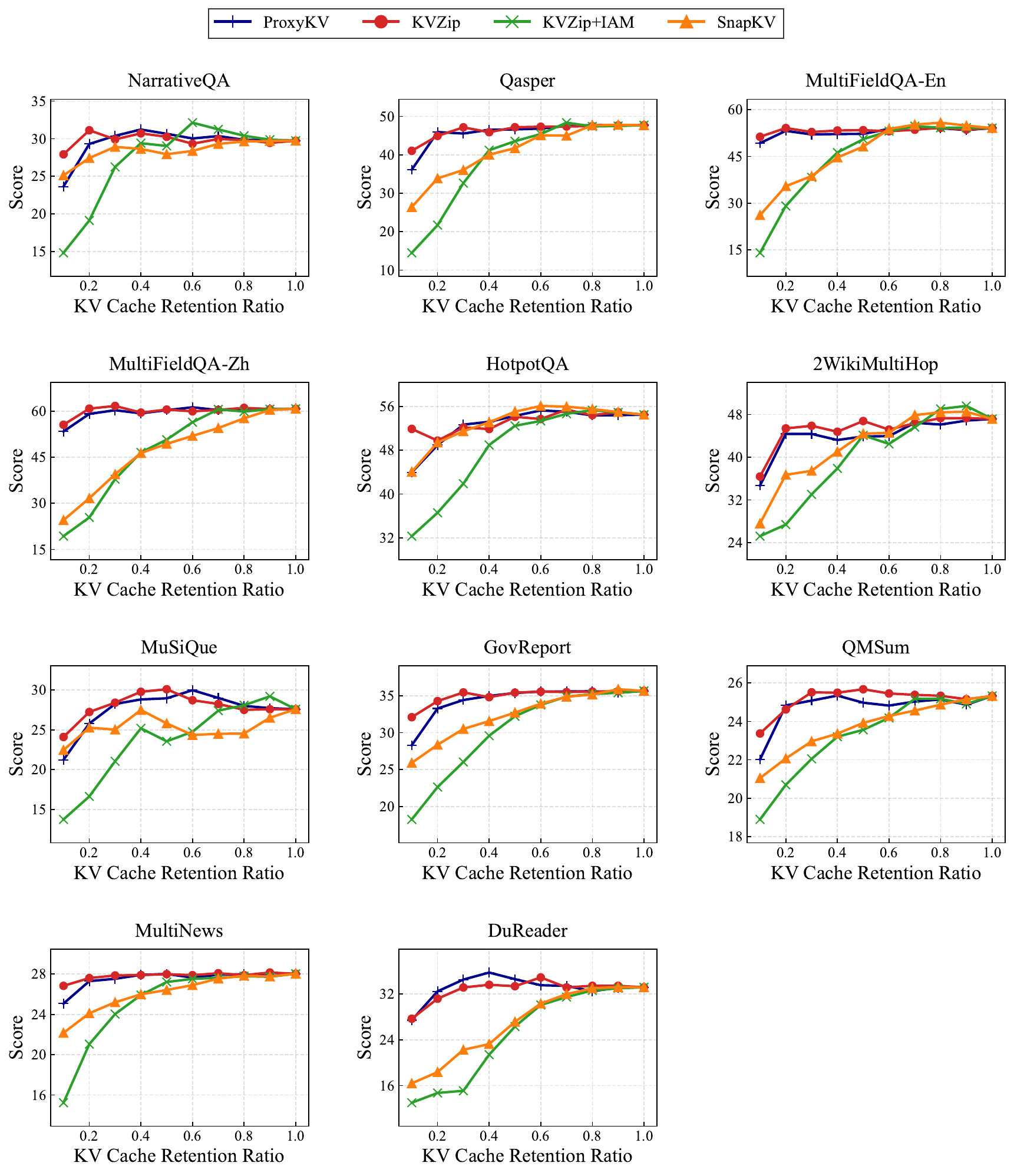}
    \caption{Complete LongBench results, Llama-3.1, Part I.}
    \label{fig:longbench_21_llama_part1}
\end{figure}

\begin{figure}[!htbp]
    \centering
    \includegraphics[width=\linewidth]{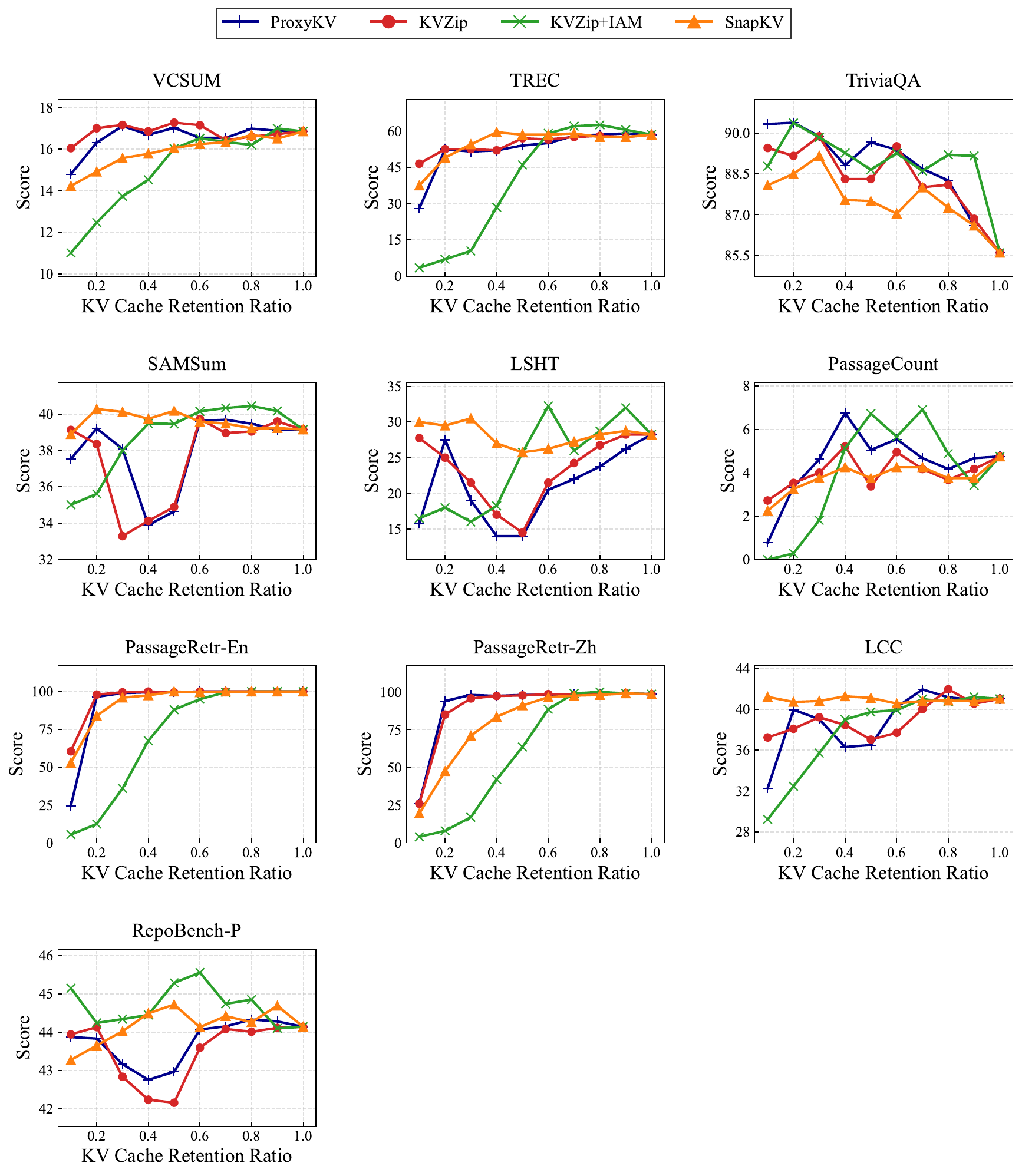}
    \caption{Complete LongBench results, Llama-3.1, Part II.}
    \label{fig:longbench_21_llama_part2}
\end{figure}

\begin{figure}[!htbp]
    \centering
    \includegraphics[width=\linewidth]{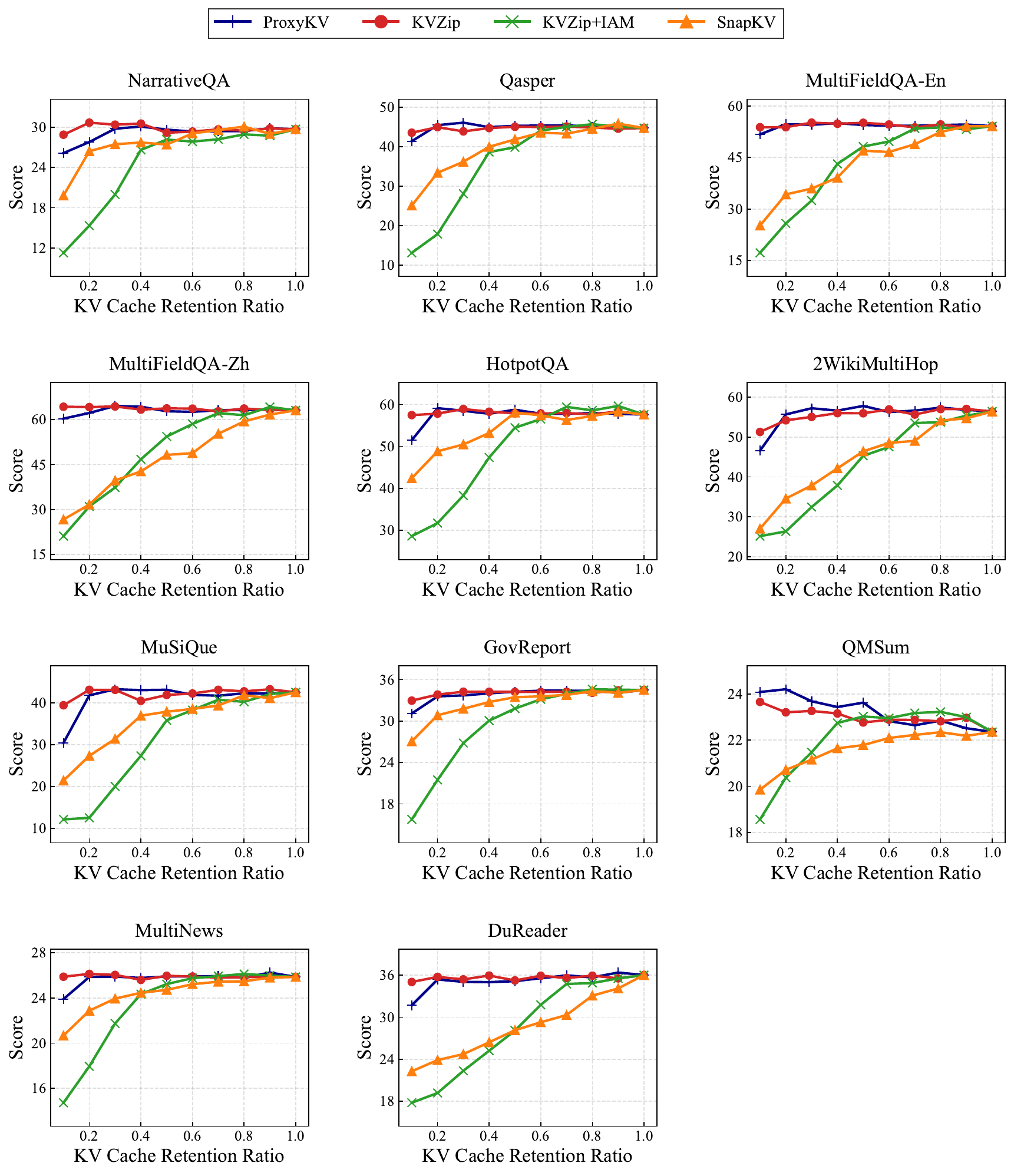}
    \caption{Complete LongBench results, Qwen-3-32B, Part I.}
    \label{fig:longbench_21_qwen332b_part1}
\end{figure}

\begin{figure}[!htbp]
    \centering
    \includegraphics[width=\linewidth]{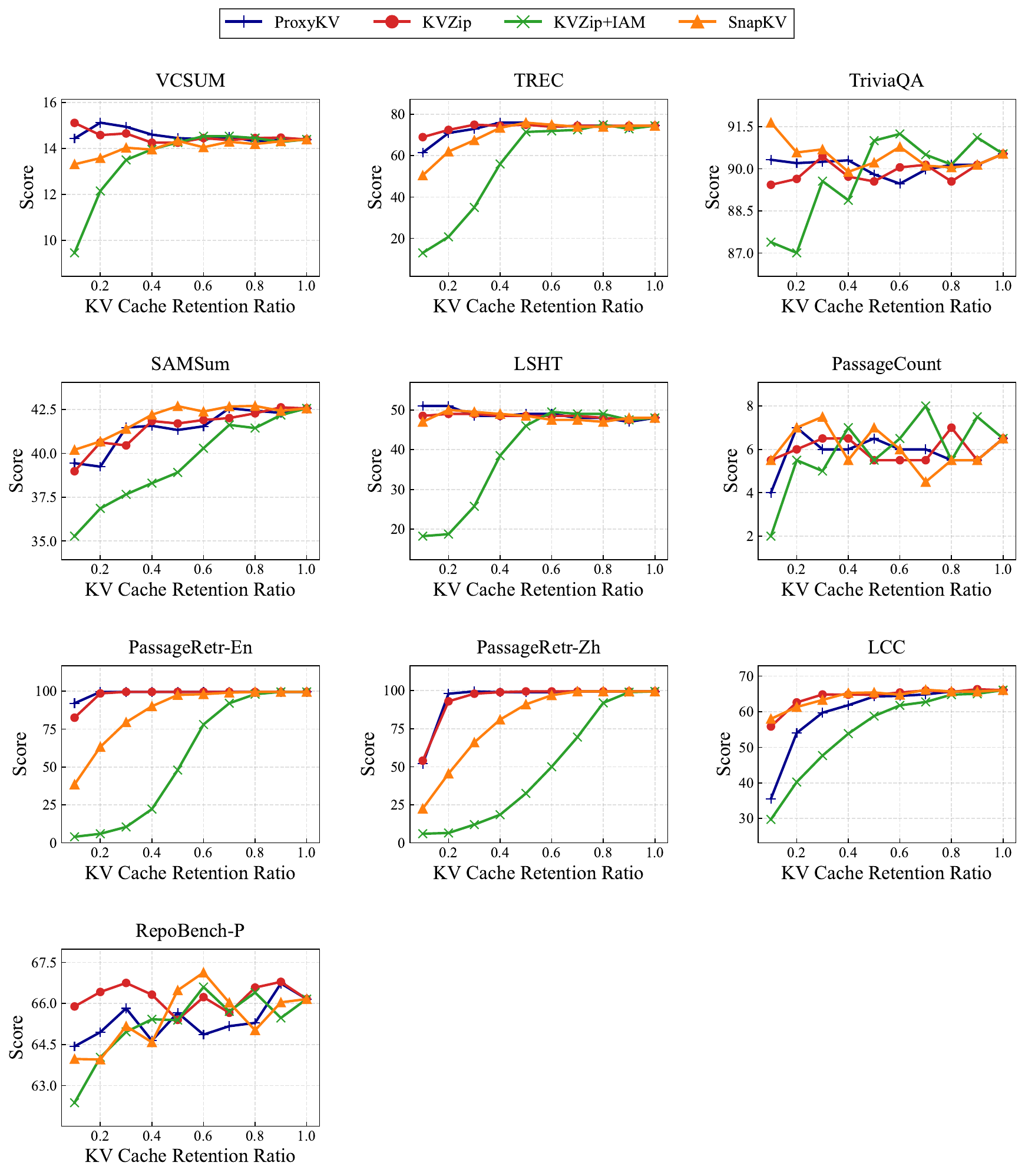}
    \caption{Complete LongBench results, Qwen-3-32B, Part II.}
    \label{fig:longbench_21_qwen332b_part2}
\end{figure}

\Cref{fig:longbench_21_qwen_part1,fig:longbench_21_qwen_part2,fig:longbench_21_llama_part1,fig:longbench_21_llama_part2,fig:longbench_21_qwen332b_part1,fig:longbench_21_qwen332b_part2} report the full 21-dataset LongBench evaluation for the Qwen-2.5-7B, Llama-3.1-8B, and Qwen-3-32B target families, using the same retention ratios and method ordering as in the main experiments. The Qwen-3-32B panels (paired with the dedicated Qwen-3-4B proxy) span a $\sim$$8\times$ target/proxy size ratio and complement the RULER stress test (\Cref{sec:ruler}) and the per-task RULER breakdown in \Cref{appendix:ruler}, which together exercise the same retention range $\rho \in \{0.1, \ldots, 0.9\}$ as LongBench while emphasizing the long-context primitives (NIAH, multi-key/value, variable tracking, word-frequency aggregation) most sensitive to scoring quality at aggressive compression.

\paragraph{Reading the panels.}
Each panel reports task accuracy as a function of the KV cache retention ratio $\rho \in \{0.1, 0.2, \ldots, 0.9\}$, with the four methods drawn in the same color and marker scheme as the main-text figures: \textbf{ProxyKV} (blue), \textbf{KVZip} (red), \textbf{KVZip+IAM} (green), and \textbf{SnapKV} (orange). Part~I covers 11 single-document QA, multi-hop reasoning, and summarization datasets where global semantic dependencies dominate; Part~II (\Cref{fig:longbench_21_qwen_part2,fig:longbench_21_llama_part2}) covers the remaining 10 datasets, predominantly retrieval, classification, and code-completion benchmarks where SnapKV's local attention spikes are most competitive, together with two additional summarization tasks (\textit{SAMSum}, \textit{VCSum}) that we group with Part~II to keep the per-page panel layout balanced.

\paragraph{Qwen-2.5-7B target.}
On the 11 retrieval/classification/code panels in Part~II, ProxyKV (blue) and KVZip (red) overlap on \emph{TREC}, \emph{TriviaQA}, \emph{LSHT}, and \emph{LCC} across all retention ratios, indicating that the proxy successfully recovers the layer-wise key signal that drives label and span retrieval. SnapKV (orange) is competitive only on \emph{PassageRetrieval-En/Zh} where the answer span is locally clustered, and KVZip+IAM (green) trails on \emph{PassageCount} and \emph{SAMSum} where the static layer alignment cannot disambiguate near-duplicate keys.

\paragraph{Llama-3.1-8B target, Part I.}
The 11 single-document QA, multi-hop, and summarization panels reproduce the cross-method ordering observed on Qwen-2.5: ProxyKV stays within $1$--$2$ points of KVZip on every panel and the gap to SnapKV widens monotonically as $\rho$ drops below $0.3$, with the largest separations on \emph{HotpotQA}, \emph{2WikiMultiHop}, and \emph{MuSiQue}, the three multi-hop tasks where multiple non-adjacent keys must be retained simultaneously and SnapKV's local-window heuristic fails first.

\paragraph{Llama-3.1-8B target, Part II.}
On the 10 retrieval/classification/code panels, ProxyKV again tracks KVZip within $1$--$2$ points across all $\rho$. The most diagnostic panels are \emph{PassageRetr-En/Zh} and \emph{LCC}: KVZip+IAM (green) lags by $20$--$40$ points at $\rho{=}0.1$--$0.3$, confirming that rigid layer-to-layer alignment is the binding constraint when the target tokens are long-tailed and unevenly distributed across layers, while ProxyKV's hybrid axial mapping closes the gap entirely.

\paragraph{Qwen-3-32B target, Part I.}
The 11 single-document QA, multi-hop, and summarization panels test whether the HybridAxialMapper recipe holds when retrained on the Qwen-3-4B proxy paired with the much larger Qwen-3-32B target ($\sim$$8\times$ target/proxy size ratio, the largest in our setup). The cross-method ordering observed on Qwen-2.5 and Llama-3.1 reproduces here without exception: ProxyKV (blue) tracks KVZip (red) within $1$--$2$ points across the entire retention range $\rho \in [0.3, 0.9]$ on all four single-doc QA panels (\emph{NarrativeQA}, \emph{Qasper}, \emph{MultiFieldQA-En/Zh}) and on every multi-hop and summarization panel, with the two curves visually overlapping above $\rho{=}0.5$. The most diagnostic panels are the multi-hop trio \emph{HotpotQA}, \emph{2WikiMultiHop}, and \emph{MuSiQue}, where multiple non-adjacent keys must be retained simultaneously: at $\rho{=}0.3$ ProxyKV matches KVZip to within $1$ point ($58.5$ vs.\ $58.9$, $57.2$ vs.\ $55.0$, and $43.3$ vs.\ $43.2$ respectively), while SnapKV (orange) trails by $5$--$11$ points and KVZip+IAM (green) collapses by $20$--$25$ points. KVZip+IAM exhibits the sharpest deterioration of the four methods on Qwen-3-32B as well: at $\rho{=}0.1$ it drops to $11.3$ on \emph{NarrativeQA}, $13.1$ on \emph{Qasper}, $17.2$ on \emph{MultiFieldQA-En}, and $14.7$ on \emph{MultiNews}, $15$--$50$ points below ProxyKV in each case, confirming that the rigid layer-to-layer alignment also fails to scale to the 32B target. The summarization panels \emph{GovReport}, \emph{QMSum}, and \emph{MultiNews} compress better than the QA panels: all four methods stay within a few points of the unpruned baseline down to $\rho{=}0.3$, since extractive summary generation tolerates aggressive eviction better than span-precise QA.

\paragraph{Qwen-3-32B target, Part II.}
The 10 retrieval, classification, and code-completion panels probe the regime where SnapKV's local attention spikes are most competitive. ProxyKV continues to track KVZip on the four classification and retrieval panels that drive the cross-scale claim: on \emph{TREC} and \emph{LSHT} the two curves overlap within $1$ point across $\rho \in [0.3, 0.9]$, on \emph{TriviaQA} all four methods saturate near $90$\% (with even SnapKV at $\rho{=}0.1$ reaching $91.6$, indistinguishable from the oracle), and \emph{PassageRetrieval-En/Zh} are the most diagnostic of all---ProxyKV preserves the unpruned $99.5$\% accuracy down to $\rho{=}0.3$ on both languages while KVZip+IAM collapses to $10.5$ (En) and $12.0$ (Zh) and SnapKV degrades to $79.5$ and $66.0$, exposing the $20$--$90$-point margin by which rigid layer alignment and local-window heuristics fail when the answer span is sparsely distributed across non-adjacent passages. At the most aggressive $\rho{=}0.1$ setting ProxyKV even surpasses the KVZip oracle on \emph{PassageRetrieval-En} ($92.0$ vs.\ $82.5$), suggesting that the proxy-driven hybrid axial mapping captures a smoother score distribution than the target's own attention under extreme compression. The one panel where ProxyKV does not dominate is \emph{LCC}: at $\rho{=}0.1$ SnapKV reaches $58.0$ versus ProxyKV's $35.6$ and KVZip's $55.9$, the only Part~II task where local-window scoring genuinely helps because code-completion targets are locally clustered around the cursor; \emph{RepoBench-P}, by contrast, remains tightly clustered ($62$--$67$) for all four methods at every $\rho$, so the LCC effect does not generalize across the code suite. The remaining panels (\emph{SAMSum}, \emph{PassageCount}, \emph{VCSum}) are non-discriminative---either saturated or near-floor for every method---so the Part~II reading at the 32B target scale extends the cross-scale conclusion of the RULER stress test (\Cref{sec:ruler}): ProxyKV recovers the KVZip oracle on the discriminative retrieval and classification primitives and never falls below the strongest baseline on any Part~II task except \emph{LCC} at the most aggressive retention.

\section{Per-task RULER breakdown}
\label{appendix:ruler}

\begin{figure}[p]
    \centering
    \includegraphics[width=\linewidth]{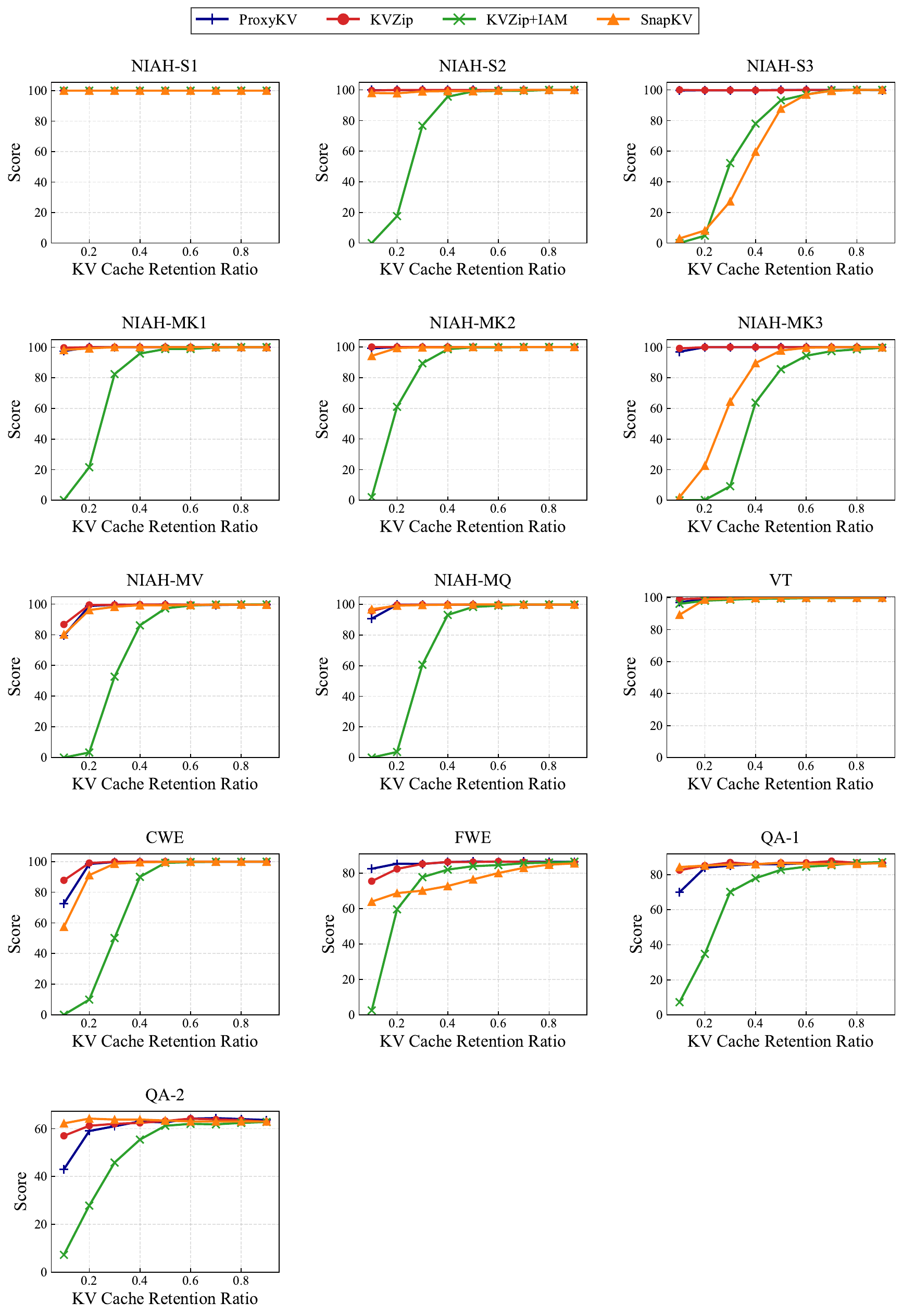}
    \caption{Complete per-task RULER results, Llama-3.1-8B target. 13 subsets $\times$ 4 methods $\times$ 9 retention ratios. Companion to the aggregate curve in \Cref{fig:ruler_avg}.}
    \label{fig:ruler_grid_llama}
\end{figure}

\begin{figure}[p]
    \centering
    \includegraphics[width=\linewidth]{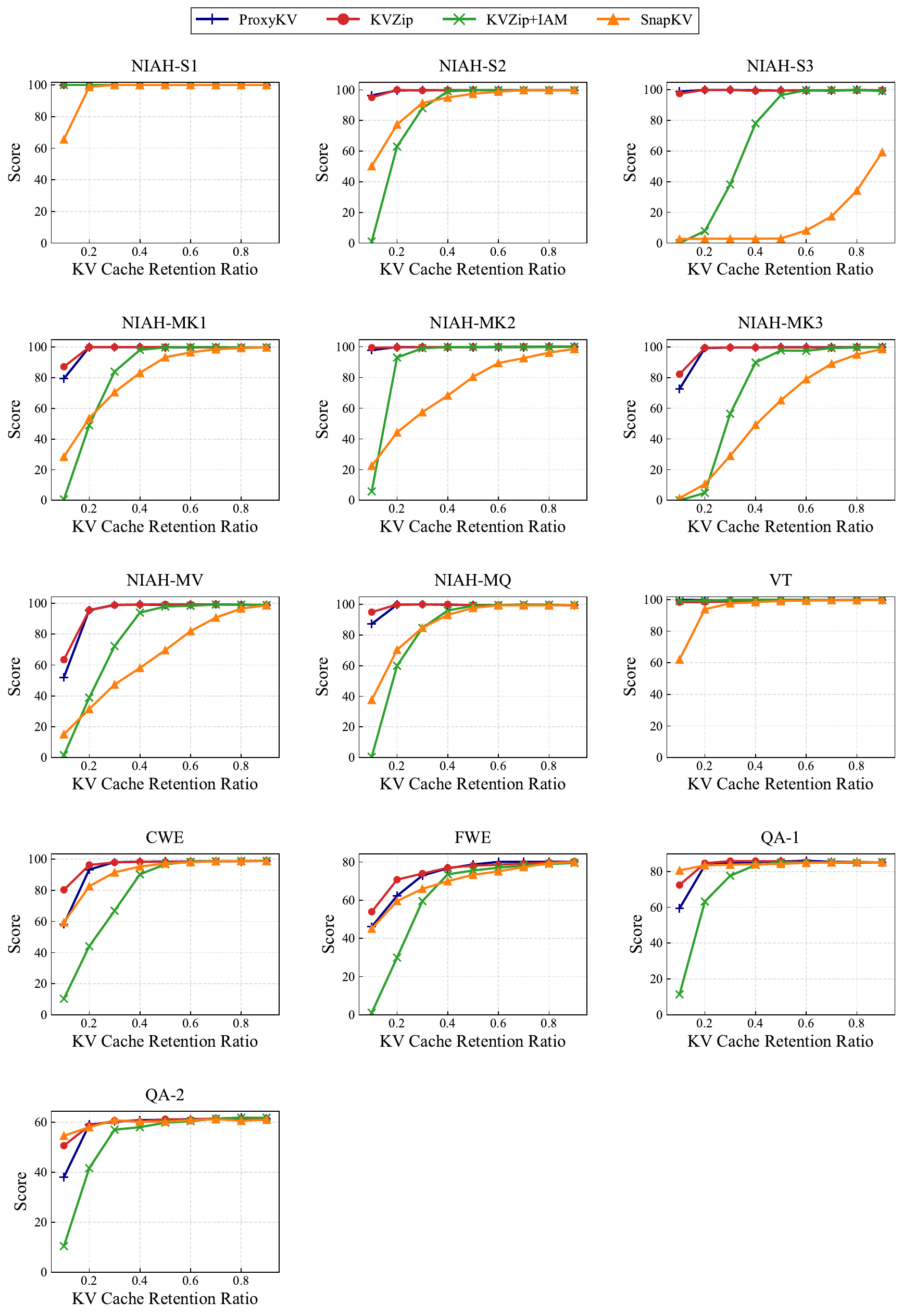}
    \caption{Complete per-task RULER results, Qwen-2.5-7B target with a Qwen-2.5-1.5B proxy. Same axes and method ordering as \Cref{fig:ruler_grid_llama}.}
    \label{fig:ruler_grid_qwen25}
\end{figure}

\begin{figure}[p]
    \centering
    \includegraphics[width=\linewidth]{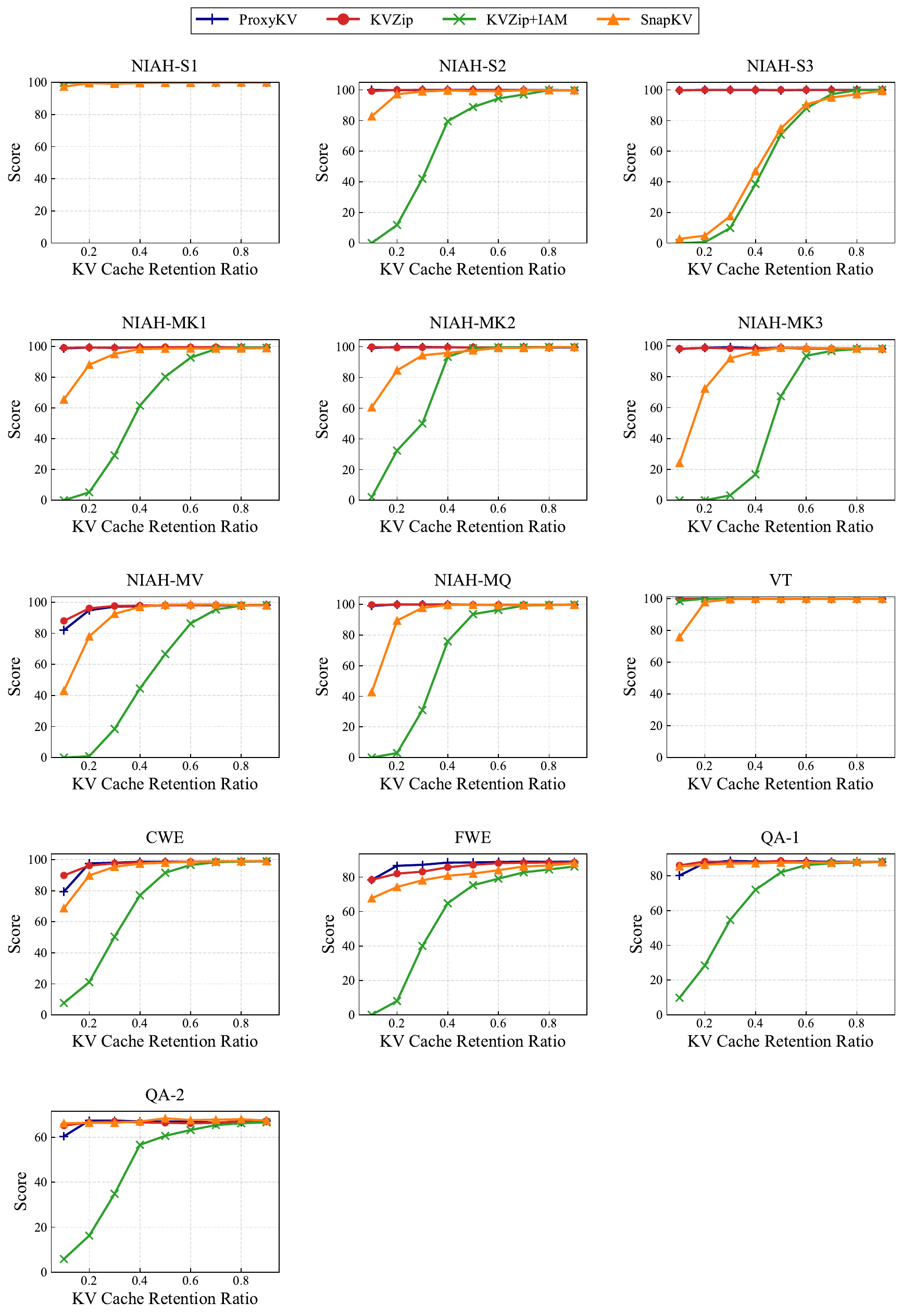}
    \caption{Complete per-task RULER results, Qwen-3-32B target paired with a dedicated Qwen-3-4B proxy ($\sim$$8{\times}$ target/proxy size ratio, the largest in our setup); ProxyKV stays within $1$--$2$ points of KVZip on every subset.}
    \label{fig:ruler_grid_qwen332b}
\end{figure}

\Cref{fig:ruler_grid_llama,fig:ruler_grid_qwen25,fig:ruler_grid_qwen332b} report the full per-task RULER breakdown for each of the three target models, expanding the aggregate curves of \Cref{fig:ruler_avg} into 13 individual panels per target. The 13 RULER subsets group into four primitive families: NIAH retrieval (single S1-S3, multi-key MK1-MK3, multi-value MV, multi-query MQ), Variable Tracking (VT), word-frequency aggregation (Common Word Extraction CWE, Frequent Word Extraction FWE), and Question Answering (QA-1, QA-2). Method coloring matches the main text: \textbf{ProxyKV} (blue), \textbf{KVZip} (red), \textbf{KVZip+IAM} (green), \textbf{SnapKV} (orange); $x$-axis is the retention ratio $\rho \in \{0.1, \ldots, 0.9\}$ and $y$-axis is task-specific accuracy.

\paragraph{Reading the panels.}
Each row collects subsets that probe a single primitive: row 1 stress-tests retrieval at increasing key-needle counts, row 2 stresses multi-value/multi-query aggregation, row 3 stress-tests cross-document tracking and word-frequency reasoning, and the final QA pair tests retrieval-conditional generation. We highlight three reading angles aligned with the main-text claims in \Cref{sec:ruler}: (i)~\textbf{NIAH multi-key/multi-value}---SnapKV (orange) collapses below $\rho{=}0.3$ where its observation-window heuristic loses the global key signal, while ProxyKV (blue) tracks the KVZip oracle (red) within $1$--$2$ points; (ii)~\textbf{CWE/FWE}---KVZip+IAM (green) trails by a wide margin, exposing the limit of static layer-to-layer alignment under long-tail token statistics; (iii)~\textbf{QA-1/QA-2}---all four methods cluster near saturation, confirming that the aggregate gap on \Cref{fig:ruler_avg} is driven by the retrieval and aggregation subsets above and not by QA. Across all 13 subsets, ProxyKV and KVZip remain visually indistinguishable on Llama-3.1-8B (Llama-3.2-1B proxy), Qwen-2.5-7B (Qwen-2.5-1.5B proxy), \emph{and} Qwen-3-32B (Qwen-3-4B proxy), demonstrating that the same HybridAxialMapper recipe holds when retrained on each intra-family pair and is not an artifact of any specific architecture or scale.

\paragraph{Llama-3.1-8B target.}
The diagnostic panel on Llama is \emph{NIAH-MK3} (multi-key with three needles): SnapKV (orange) collapses below $\rho{=}0.3$ because its observation-window scoring loses the global key signal once aggressive eviction begins, while ProxyKV (blue) tracks KVZip (red) to within $1$--$2$ points down to $\rho{=}0.1$. \emph{CWE} and \emph{FWE} surface the limit of static layer alignment: KVZip+IAM (green) drops by $20$--$30$ points relative to ProxyKV across the entire $\rho$ range, confirming that the long-tail token distribution requires the per-layer adaptive routing learned by the HybridAxialMapper. The \emph{QA-1}/\emph{QA-2} panels saturate near the upper bound for all four methods, so the aggregate gap reported in \Cref{fig:ruler_avg} is driven almost entirely by retrieval and aggregation primitives, not by question answering.

\paragraph{Qwen-2.5-7B target.}
On the Qwen-2.5-7B target paired with the Qwen-2.5-1.5B proxy, the two curves are visually indistinguishable on 11 of the 13 subsets, with the only marginal separation on \emph{NIAH-S3} at $\rho{=}0.1$. The relative ordering of methods is otherwise identical to Llama: SnapKV remains the weakest baseline on every multi-needle and multi-value subset, and KVZip+IAM exhibits the same long-tail collapse on \emph{CWE}/\emph{FWE}. The fact that the per-family Qwen-2.5 result is not visibly better than the per-family Llama result rules out the concern that ProxyKV's accuracy is dominated by any one family's structural prior --- the same HybridAxialMapper recipe, separately retrained per intra-family pair, recovers the KVZip oracle on both architectures.

\paragraph{Qwen-3-32B target (largest target/proxy size ratio).}
Qwen-3-32B is the most stringent target in our setup: paired with the Qwen-3-4B proxy, the target/proxy size ratio reaches $\sim$$8{\times}$, the widest gap we evaluate. Despite this, ProxyKV (blue) overlays the KVZip oracle (red) on every one of the 13 panels with no visible separation at any $\rho$, including the previously discriminating \emph{NIAH-MK3}, \emph{NIAH-MV}, and \emph{CWE}/\emph{FWE} primitives. The relative ordering of the two baselines, however, \emph{flips} at this scale: KVZip+IAM (green) is now the weakest method on the aggregate (\Cref{fig:ruler_avg}) and exhibits the most severe collapse of any panel in our entire study---dropping to $0.0$ on \emph{NIAH-MK3}, \emph{NIAH-MV}, and \emph{FWE} at $\rho{=}0.1$ and recovering only past $\rho{=}0.5$ on the multi-needle and aggregation primitives---so the static layer-to-layer alignment becomes \emph{less}, not more, viable at the 32B target. SnapKV (orange) is correspondingly the stronger of the two baselines here: at $\rho{=}0.3$ it stays at $84$--$96$ on the multi-key panels (\emph{MK1}/\emph{MK2}/\emph{MK3}) where KVZip+IAM is at $29$/$50$/$3$, and on the 13-task average at $\rho{=}0.3$ SnapKV reaches ${\approx}85$ versus KVZip+IAM's ${\approx}43$. The takeaway is that the per-layer adaptive routing learned by the HybridAxialMapper does \emph{not} commute with target scale---the same intra-family pair retraining recipe that makes ProxyKV scale-invariant also exposes the rigidity of static alignment more sharply at 32B. This figure is the central empirical evidence for the scale-up claim of \Cref{sec:ruler}: ProxyKV is the only method that recovers the oracle at the 32B target scale.

\section{Implementation details and training protocols}
\label{appendix:training_settings}

To map the KV cache saliency from the target model to the proxy model, we employ a multi-objective training protocol. The specific configurations are detailed below.

\paragraph{Optimization Strategy.}
The HybridAxialMapper is trained for 30 epochs using the AdamW optimizer with a learning rate of $2\times 10^{-4}$ and a weight decay of $1\times 10^{-4}$. We implement a linear warmup for the first $1{,}000$ steps, followed by a \texttt{ReduceLROnPlateau} scheduler with a patience of 3 epochs and a decay factor of $0.5$. We use a batch size of 8 and reserve $5\%$ of the samples for validation. The two mappers for the Llama-3.1-8B and Qwen-2.5-7B targets are each trained on a single NVIDIA RTX PRO 6000 GPU (the same hardware used for evaluation in \Cref{sec:efficiency}) for approximately 14 hours; the mapper for the Qwen-3-32B target is trained on 4$\times$ NVIDIA A100 GPUs for approximately 10 hours. Total mapper-training compute is approximately $68$ GPU-hours; counting preliminary and ablation runs, the full study consumed on the order of a few hundred GPU-hours. The trained mapper checkpoints are lightweight: approximately $100$\,MB on disk for the Llama-3.1-8B and Qwen-2.5-7B mappers, and approximately $150$\,MB for the Qwen-3-32B mapper, several orders of magnitude smaller than their respective target models.

\paragraph{Multi-Objective Loss.}
The training objective is the composite loss $\mathcal{L} = \sum_i \lambda_i \mathcal{L}_i$. The loss coefficients are empirically set to $\lambda_{bin}{=}10.0$, $\lambda_{mse}{=}20.0$, $\lambda_{fine}{=}3.0$, $\lambda_{global}{=}2.0$, and $\lambda_{cos}{=}0.5$. The Multi-Ratio Binary Loss ($\mathcal{L}_{bin}$) evaluates retention ratios $\rho \in \{0.05, 0.1, 0.15, 0.2, 0.3, 0.4, 0.5\}$ using a power-law decay to emphasize high-compression regimes. We additionally use Value-Weighted MSE ($\alpha{=}1.5$) and Hierarchical Ranking losses to ensure numerical alignment and preserve the relative importance of attention scores.

\paragraph{Data Mixture and Augmentation.}
Training data is curated from a diverse mixture of benchmarks, including GSM8K, SQuAD, NIAH, and multiple SCBench subsets (\textit{QA ENG}, \textit{Summary}, \textit{KV}, \textit{Many Shot}, \textit{Choice Eng}, and \textit{Prefix Suffix}). To accommodate long-context inputs, we process sequences using a sliding window strategy with a crop length of $2{,}048$ tokens and a stride of $1{,}024$.

\begin{figure}[!t]
    \centering
    \includegraphics[width=0.95\linewidth]{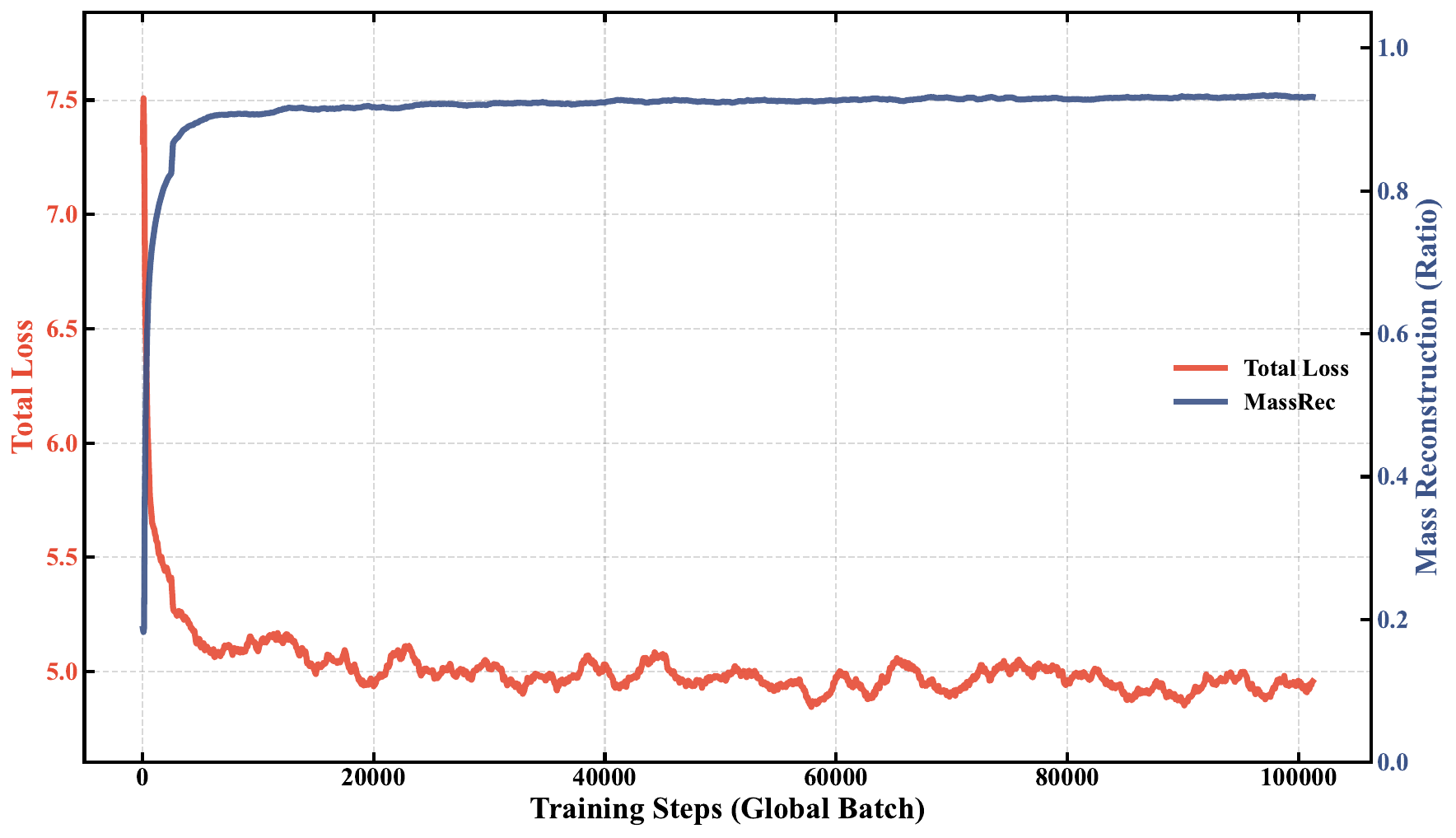}
    \caption{Mass Reconstruction stabilizes near $0.95$ within the first quarter of training, confirming that the Multi-Granularity Hybrid Loss converges smoothly without overfitting. Total loss (red) and Mass Reconstruction ratio (blue) over $100{,}000$ training steps.}
    \label{fig:training_dynamics}
\end{figure}

\paragraph{Training Dynamics.}
\Cref{fig:training_dynamics} illustrates the convergence of the total loss (red) alongside the Mass Reconstruction ratio (blue) over $100{,}000$ global training steps. The consistent decline in total loss and the rapid stabilization of the Mass Reconstruction ratio near $0.95$ demonstrate the effectiveness of our Multi-Granularity Hybrid Loss in distilling high-fidelity importance scores. The smooth convergence further confirms that the mapper successfully captures structural attention invariants without overfitting to the specific linguistic patterns of the training mixture. \emph{Note that the Mass Reconstruction ratio (${\approx}0.95$) is an intra-training metric measuring how much of the target's attention probability mass is captured by the predicted Top-$K$ set, and is not directly comparable to the end-to-end task-accuracy recovery (${\approx}98.7\%$) reported in the main experiments.}

\end{document}